\documentclass[lettersize,journal]{IEEEtran}
\usepackage{amsmath,amsfonts}
\usepackage{algorithmic}
\usepackage{algorithm}
\usepackage{array}
\usepackage[caption=false,font=normalsize,labelfont=sf,textfont=sf]{subfig}
\usepackage{textcomp}
\usepackage{stfloats}
\usepackage{url}
\usepackage{verbatim}
\usepackage{graphicx}
\usepackage{cite}
\usepackage{adjustbox}

\def\BibTeX{{\rm B\kern-.05em{\sc i\kern-.025em b}\kern-.08em
    T\kern-.1667em\lower.7ex\hbox{E}\kern-.125emX}}
\usepackage{balance}

\usepackage{placeins}

\usepackage{amssymb}

\usepackage[pagebackref,breaklinks,colorlinks]{hyperref}

\usepackage{float} 

\usepackage{bbm}
\usepackage{hhline}
\usepackage{boldline}
\usepackage{multirow}
\usepackage{pifont}
\usepackage{xcolor}
\usepackage{booktabs}
\usepackage{makecell}
\usepackage{wrapfig}
\usepackage{orcidlink}

\usepackage[table]{xcolor}
\usepackage{colortbl}

\def\eg{\emph{e.g.},\ } 
\def\ie{\emph{i.e.},\ }

\usepackage{xspace}
\makeatletter
\DeclareRobustCommand\onedot{\futurelet\@let@token\@onedot}
\def\@onedot{\ifx\@let@token.\else.\null\fi\xspace}

\AtEndPreamble{
    \usepackage[capitalize]{cleveref}
    \crefname{section}{Sec.}{Secs.}
    \Crefname{section}{Section}{Sections}
    \Crefname{table}{Table}{Tables}
    \crefname{table}{Tab.}{Tabs.}
}

\begin{document}

\title{GOT-JEPA: Generic Object Tracking with Model Adaptation and Occlusion Handling using Joint-Embedding Predictive Architecture}

\author{Shih-Fang Chen\orcidlink{0000-0002-8438-400X}, 
Jun-Cheng Chen\orcidlink{0000-0002-0209-8932}~\IEEEmembership{Member,~IEEE}, 
I-Hong Jhuo\orcidlink{0009-0009-3893-3758}~\IEEEmembership{Member,~IEEE},
\\and Yen-Yu Lin\orcidlink{0000-0002-7183-6070}~\IEEEmembership{Senior Member,~IEEE}


\thanks{
Shih-Fang Chen and Yen-Yu Lin are with the Department of Computer Science, National Yang Ming Chiao Tung University (e-mail: csf.cs09@nycu.edu.tw; lin@cs.nycu.edu.tw) (Corresponding author: Y-Y Lin). 
}  
\thanks{
Jun-Cheng Chen is with the Research Center for Information Technology Innovation, Academia Sinica (e-mail: pullpull@citi.sinica.edu.tw). 
} 
\thanks{
I-Hong Jhuo is with Microsoft AI (e-mail: ihjhuo@gmail.com). 
}
\thanks{
Copyright © 2026 IEEE. Personal use of this material is permitted. However, permission to use this material for any other purposes must be obtained from the IEEE by sending an email to pubs-permissions@ieee.org.
}

}

\markboth{IEEE TRANSACTIONS ON CIRCUITS AND SYSTEMS FOR VIDEO TECHNOLOGY}%
{Shell \MakeLowercase{\textit{et al.}}: A Sample Article Using IEEEtran.cls for IEEE Journals}




\maketitle



\begin{abstract}

The human visual system tracks objects by integrating current observations with previously observed information, adapting to target and scene changes, and reasoning about occlusion at fine granularity. In contrast, recent generic object trackers are often optimized for training targets, which limits robustness and generalization in unseen scenarios, and their occlusion reasoning remains coarse, lacking detailed modeling of occlusion patterns. To address these limitations in generalization and occlusion perception, we propose GOT-JEPA, a model-predictive pre-training framework that extends JEPA from predicting image features to predicting tracking models.
Given identical historical information, a teacher predictor generates pseudo-tracking models from a clean current frame, and a student predictor learns to predict the same pseudo-tracking models from a corrupted version of the current frame. This design provides stable pseudo supervision and explicitly trains the predictor to produce reliable tracking models under occlusions, distractors, and other adverse observations, improving generalization to dynamic environments.
Building on GOT-JEPA, we further propose OccuSolver to enhance occlusion perception for object tracking. OccuSolver adapts a point-centric point tracker for object-aware visibility estimation and detailed occlusion-pattern capture. Conditioned on object priors iteratively generated by the tracker, OccuSolver incrementally refines visibility states, strengthens occlusion handling, and produces higher-quality reference labels that progressively improve subsequent model predictions.
Extensive evaluations on seven benchmarks show that our method effectively enhances tracker generalization and robustness. 
The code will be available at~\url{https://github.com/chenshihfang/GOT}.

\vspace{0.1 in}


\end{abstract}


\section{Introduction}
\label{sec:intro}


%
\IEEEPARstart{G}{eneric} Object Tracking (GOT)~\cite{DiMP, SiamRPN++, jav_got_survey} allows machine visual tracking of any arbitrary target object specified by only an initial bounding box in the first frame and predicting its location in subsequent frames.
Given this limited information, learning a robust tracker to accurately predict the target location in each frame of a dynamic environment is very challenging, especially in adverse conditions, where unseen target objects, complex distractors, and object deformations are present.
Additionally, handling occlusions and out-of-view scenarios is crucial for long-term tracking.

%


\begin{figure}[!t] \centering
\includegraphics[width=0.49\textwidth]{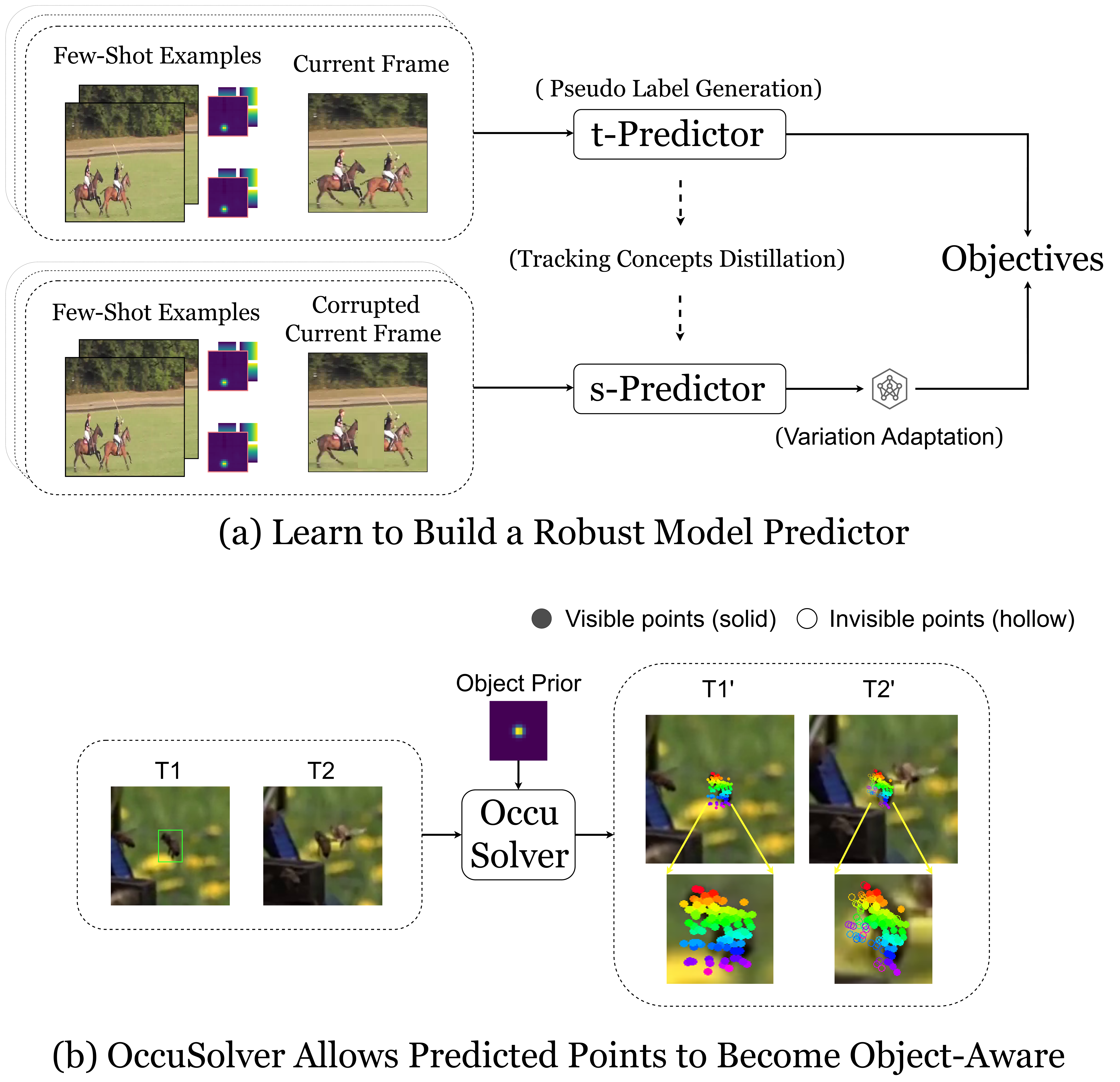}
\vspace{-0.20in}
%
\caption{(a) GOT-JEPA extends the JEPA architecture to build a robust model predictor for model adaptation.
The t-Predictor generates diverse tracking models for pseudo-labeling, while the s-Predictor predicts them using corrupted current frames.
Both predictors use identical historical frames and tracking results as few-shot examples to aid target identification, while variations in the current frame drive robustness in tracking-model prediction.
(b) OccuSolver enhances occlusion-handling capabilities. 
T1' shows initial point sampling.
T2' shows that OccuSolver filters redundant points as invisible states, retaining essential points as visible states.
This information is then utilized to enhance occlusion perception in the proposed GOT tracker.
}
\label{fig:FM_T-teaser}
\vspace{-0.2in}
\end{figure}

%

The tracking-by-detection paradigm~\cite{jav_got_survey} is widely used to learn trackers capable of handling dynamically changing targets in adverse scenarios~\cite{DiMP, ToMP}.
These trackers typically consist of a {\em model predictor} and a {\em localization head}.
The model predictor dynamically updates the 
tracker so that the localization head of the updated tracker can better identify the target in the coming frames.
To further address challenges such as occlusions and out-of-view scenarios, these trackers may estimate confidence scores and dynamically update their estimates.
However, confidence scores are typically derived from appearance-based non-occlusion perception, whereas occlusion labels exhibit deficiencies in GOT, leading to unreliable confidence scores, especially when occlusions are present.
The existing tracking-by-detection paradigm thus suffers from limited generalization and occlusion-handling capabilities.

Specifically, the model predictor of existing methods\cite{ToMP,PiVOT} is prone to generating less reliable trackers when inferring targets unseen during training. 
This is because the existing paradigm optimizes the model predictor during training to best identify learned tracking targets, rather than learning model prediction as a general skill.
This limited learning paradigm hinders the model predictor's ability to adapt to unseen targets.
To mitigate challenges such as occlusion and target disappearance, existing trackers have explored several directions. One line of research maintains multiple trajectories to track plausible targets and improve confidence reliability~\cite{DeepMTA,ARTrackV2}. 
Another line enhances occlusion robustness by promoting appearance invariance through masking strategies (e.g., masked autoencoders)~\cite{SiamON, ORTrack}. 
Alternative methods estimate target locations under low-confidence predictions to assist occlusion recovery~\cite{QRDT} or incorporate visibility-aware confidence estimation~\cite{ToMP,PiVOT}. 
However, these methods share a common limitation in occlusion handling: they operate at the scene or box level and do not explicitly infer which target regions remain visible under partial occlusion. This limitation is further compounded by the scarcity of fine-grained occlusion annotations in generic object tracking, which leads to largely implicit supervision and limits the learning of detailed occlusion patterns.

Human cognition tracks objects by continuously integrating current observations with past information, adapting to changes in targets and environments, and reasoning about partial occlusion and visibility at fine granularity, even for unseen targets. In contrast, current object-tracking systems lack the abstract reasoning capabilities needed to handle complex occlusions and unseen targets in dynamic environments.
To bridge this gap, we introduce GOT-JEPA, as shown in~\cref{fig:FM_T-teaser}. Our work advances the JEPA (Joint-Embedding Predictive Architecture) paradigm~\cite{JEPA} from image feature prediction to the novel task of tracking model prediction. 
In the model-predictive learning paradigm of GOT-JEPA, a teacher predictor generates pseudo-tracking models from a clean current frame, while a student predictor learns to predict these models from a corrupted current frame, with identical historical conditioning for both predictors. This objective encourages recovery of target–background discrimination under degraded observations, yielding a robust and discriminative model predictor that generalizes to unseen objects and diverse environments.

The proposed OccuSolver enhances occlusion perception in GOT by combining GOT-derived object priors with the point tracker~\cite{Cotracker}, thereby integrating high-level object semantics with low-level point visibility.
Since the point tracker is point-centric and relies on initial query points from the object bounding box in GOT, these points may come from both the target and the background, leading to uncertainty about which points are beneficial to GOT. 
We thus align the point tracker with GOT by refining its output using GOT-derived object priors and objectives.
The object priors further guide dynamic point sampling, adapting to new viewpoints and surfaces not covered by the initial sampling.
OccuSolver then estimates the point-wise visibility state of a tracked object and enables the JEPA-trained model predictor to generate more effective target-adaptive tracking models by accounting for point-wise visibility states.
It turns out that a tight coupling between GOT and OccuSolver is created:
The tracker provides OccuSolver with improved object priors to identify point visibility, while the identified visibility states, in turn, help generate better reference labels for tracking model adaptation.

The main contributions are summarized as follows:
First, we introduce GOT-JEPA, a model-predictive learning framework that extends the JEPA paradigm from image-feature prediction to tracking-model prediction for streaming inputs. GOT-JEPA trains a model predictor in which a teacher predictor produces pseudo-tracking models from clean frames, and a student predictor learns to predict them from corrupted frames, with identical historical conditioning for both predictors. This design improves robustness and generalization to unseen targets and dynamic scene variations.
Second, we propose OccuSolver, which equips generic object tracking with fine-grained occlusion reasoning by tightly integrating high-level semantic perception with low-level geometric cues. Specifically, OccuSolver makes a point tracker object-aware by incorporating GOT-derived object priors and transferring point-level visibility signals to GOT-JEPA.
These target-centric visibility cues enhance occlusion reasoning and provide higher-quality reference labels for tracking-model adaptation, stabilizing subsequent model predictions, and improving recovery after reappearance.
Extensive experiments on seven benchmarks show consistent gains under occlusion and deformation, with superior generalization to both in-distribution and out-of-distribution targets.
\section{Related Work}
\label{sec:related}

In this section, we discuss several research topics relevant to the development of our method. 
\vspace{-0.1 in}
\begin{figure*}[!ht]
	\centering
        \includegraphics[width=0.98\linewidth, 
        trim={0cm 0 0cm 0}, clip]{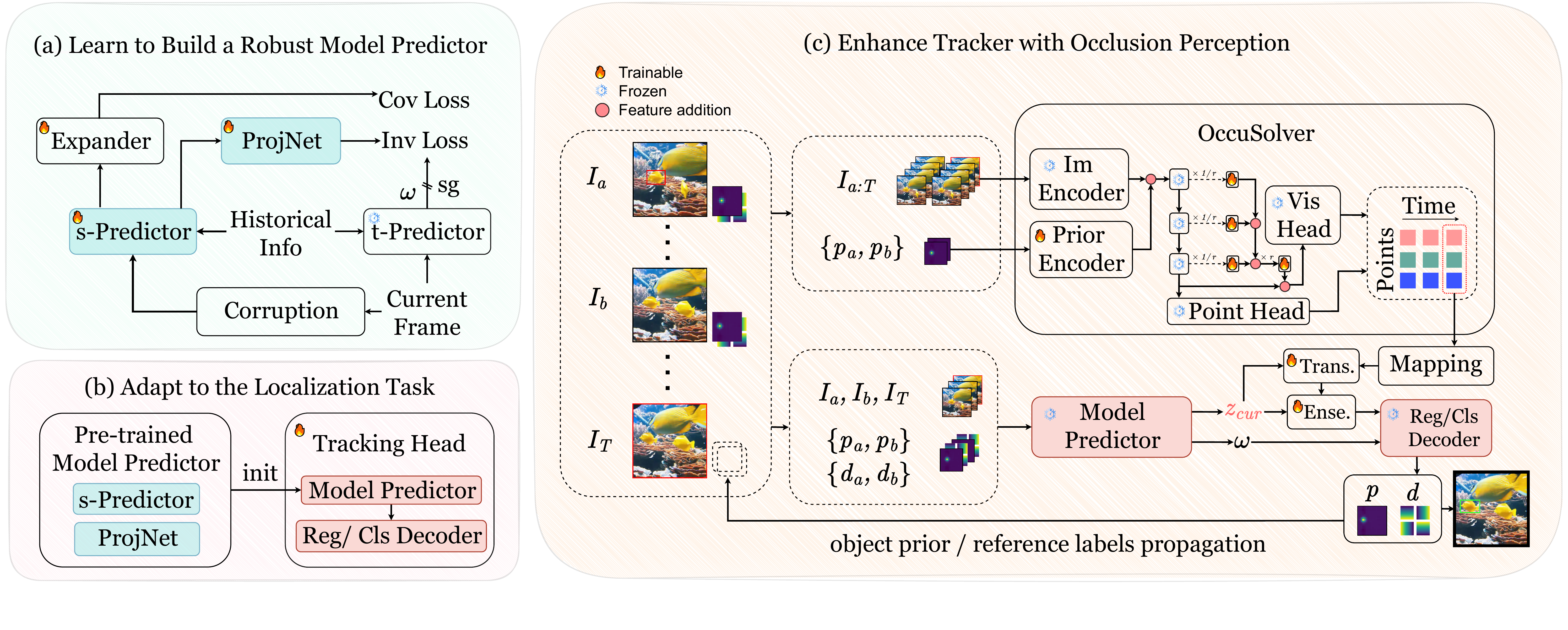}
	\vspace{-0.4 cm}
        %
	\caption{\textbf{Overview of the proposed framework.}
        (a) We pre-train a robust model predictor using a JEPA-based approach. Conditioned on identical past information, a student predictor (s-Predictor) learns from a corrupted current frame to predict the tracking models generated by a teacher (t-Predictor) with the uncorrupted input. This process compels the student to learn representations that are robust to frame variations. Details are provided in~\cref{sec:GOT-JEPA}. 
        (b) The pre-trained student predictor is then integrated into the tracking head with classification and regression decoders and fine-tuned for precise object localization. Details are provided in~\cref{sec:Background} and~\cref{sec:GOT-JEPA}. 
        (c) To address occlusions, OccuSolver adapts a point tracker to be object-aware using priors from the object tracker. The resulting point visibility states are then integrated with visual features via an Ensemble Network, enabling the final model to better handle occluded targets and generate more accurate tracking models over time. Refer to~\cref{sec:OccuSolver} for details.
        }
	\label{fig:main_figure}
\end{figure*}






\subsection{Tracking-by-Detection}

Some trackers, such as~\cite{dong2016occlusion, TCSVT_8387770, TIP_9107463, TLD, RTracker,Henriques14, RTINet,BACFgaloogahi,lukezic2017discriminative, bolme2010visual,hen_exploiting,ECO,danelljan2016beyond,liang2024joint,TIP_8662649,TrDiMP}, approach GOT as a tracking-by-detection task~\cite{jav_got_survey}. 
Recent trackers~\cite{ToMP, DiMP} following this paradigm utilize paired reference images and labels alongside the current frame to generate a tracking model in the form of discriminative correlation filters. 
This tracking model is produced through a predictor derived using meta-learning techniques~\cite{MAML, meta_Hypernetworks, SNAIL}.
A classification and regression head is also included to localize the target in the current frame.
Despite the meticulous architecture design, their model predictors still tend to produce unreliable tracking models 
because they are prone to guiding the model predictor to derive tracking models for familiar targets during training, limiting generalization capability to unseen scenarios.
In contrast, we propose GOT-JEPA, which learns to predict trackers using diverse pseudo-tracking models and effectively generalizes to both in- and out-of-distribution targets.


\subsection{Matching-based Tracking}

Several trackers~\cite{SiamFC, SiamRPN++, SINT, bao2023siamthn,shen2024overlapped,zhu2024srrt,zhang2024probabilistic,gao2023online,dong2025feature,BAYN} formulate GOT as similarity learning followed by matching~\cite{jav_got_survey}.
They focus on models capable of discriminating and matching templates within the current frame.
A deep network trained offline to learn a matching function between templates and search images is later deployed online for tracking.
This paradigm has evolved from conventional CNN-based trackers~\cite{SiamFC, SiamFC++, SiamBAN, SiamRPN, SiamCAR, Siam_R-CNN, SiamAttN, Ocean,SparseTT,RBO,CIA,IDSTT,chen2025linr,hu2023transformer,chen2024local}, 
to recent transformer-based trackers~\cite{STARK, TransT, OSTrack, GRM, ROMTrack, TransInMo, AiATrack,TATrack,RFGM,CiteTracker,MATrack,OSTrack-Aug,yan2019skimming,IDSTT,GOLA}.
Despite the improvements offered by Transformers in learning more discriminative matching functions or adopting advanced paradigms, such as auto-regressive~\cite{SeqTrack, ARTrack, AQATrack}, visual prompting~\cite{EVPTrack, HIPTrack}, masked image modeling~\cite{CTTrack,DropMAE,MAT,Context-Guided}, and diffusion~\cite{Diff-Tracker,DiffusionTrack}, these trackers exhibit limitations in generalization, compared to aforementioned model-prediction-based trackers, due to the inherent constraints of offline model training.


\subsection{Occlusion Handling for Trackers}

Researchers have developed various recovery and robustness strategies. Methods such as GlobalTrack~\cite{Globaltrack}, DeepMTA~\cite{DeepMTA}, and ARTrackV2~\cite{ARTrackV2} perform global searches or maintain multiple candidate trajectories to re-detect targets after they disappear, while ensuring temporal trajectory consistency. SiamON~\cite{SiamON} and ORTrack~\cite{ORTrack} employ masking techniques or synthetic occlusion generation to encourage the model to learn features that persist when parts of the object are hidden. Trackers such as ToMP~\cite{ToMP} and PiVOT~\cite{PiVOT} use confidence score maps to detect object visibility and mitigate occlusion.

However, these approaches predominantly operate at a coarse granularity, treating the target as a single bounding-box entity. They typically rely on global confidence scores and cannot distinguish which specific parts of an object are visible or hidden during partial occlusion. In contrast, our method integrates the proposed OccluSolver to enable fine-grained, pixel-level occlusion perception.
By explicitly estimating the visibility of discrete physical points using a point tracker~\cite{Cotracker} and applying our proposed point-tracker adaptation method to make these cues object-tracking-aware, our approach improves the object tracker's ability to distinguish occluded and unoccluded target sub-regions and maintain high precision even in complex scenarios.


\subsection{Point Trackers}
Point tracking introduced in TAP-Vid~\cite{TAP-Vid} aims to track arbitrary physical points on surfaces over a video. 
As TAP-Vid exhibits limitations in handling occlusions, in response, PIPs~\cite{PIPs} and TAPIR~\cite{TAPIR} present point-tracking models that handle occlusions.
These methods track points independently. 
OmniMotion~\cite{OmniMotion} addresses this issue, but its test-time optimization is computationally inefficient.
Cotracker~\cite{Cotracker} subsequently presents an efficient algorithm for tracking multiple points, while maintaining their correspondences.
We modify Cotracker and integrate it into our OccuSolver to facilitate GOT.
In contrast to the original Cotracker, our modified point tracker can leverage object priors provided by GOT, aligning its learning objectives with those of GOT.
This enables the point tracker's knowledge to be adapted, helping GOT better handle partially occluded regions of the target object.



\subsection{Joint-Embedding Predictive Architecture (JEPA)}
JEPA~\cite{JEPA} aims to predict the embedding of one signal from a compatible context signal. By performing prediction in representation space, JEPA can reduce reliance on irrelevant details and enable learning of transferable representations that improve transfer to downstream tasks.
Building on this paradigm, I-JEPA~\cite{I-JEPA} and V-JEPA~\cite{V-JEPA} infer missing content in image representation space from corrupted inputs, emphasizing semantic cues over pixel-level details. S-JEPA~\cite{S-JEPA} applies JEPA to 3D skeletons for action recognition, and BrainJEPA~\cite{BrainJEPA} extends JEPA to modeling of brain dynamics.

We adapt JEPA to tracking by shifting the traditional prediction target from semantic image representations to our proposed target-conditioned, discriminative tracking models. In our setting, the tracking model captures target–background discrimination conditioned on the observed history and the current frame. We train the model predictor using the proposed GOT-JEPA learning paradigm: a teacher predictor generates pseudo-tracking models from an uncorrupted current frame, and a student predictor learns to predict them from a corrupted version, with identical historical conditioning for both predictors. This objective promotes consistent tracking-model prediction under diverse frame perturbations induced by data augmentation, encouraging a robust and discriminative model predictor that generalizes beyond target-specific optimization.


\section{Method}
\label{sec:method}

In this work, we propose to enhance the generalization and robustness of a tracker through GOT-JEPA and OccuSolver. 
An overview of our method is illustrated in~\cref{fig:main_figure}.
Our tracker enhances the generalization capability by learning to predict tracking models within the proposed GOT-JEPA pre-training paradigm.
OccuSolver further utilises object priors from GOT to adapt a point tracker, thereby enhancing its alignment with GOT for improved occlusion handling and enhanced tracking model prediction.
We introduce the background of the tracking-by-detection paradigm in~\cref{sec:Background}.
Then, we describe our GOT-JEPA in~\cref{sec:GOT-JEPA} and
OccuSolver in~\cref{sec:OccuSolver}, respectively.
%

\subsection{Background}
\label{sec:Background}
In this work, we focus on tracking-by-detection-based trackers~\cite{ToMP,PiVOT,DiMP, jav_got_survey} due to their enhanced adaptability.
These trackers generally comprise an image encoder and a Tracking Head that consists of a model predictor, a regression decoder (RegDec), and a classification decoder (ClsDec).

The model predictor learns to predict the tracking models (or filters)
which are used to detect and localize the target in the incoming frames. 
Specifically, the model predictor in~\cite{PiVOT,ToMP} processes the encoded features of 
the reference frames,
the reference labels 
of the reference frames, 
and the current frame 
as inputs, where a frame is of feature resolution $H \times W$,
label elements $p_a, p_b \in \mathbb{R}^{H \times W}$ are classification score map labels, and label elements $\mathit{d_a}, \mathit{d_b} \in \mathbb{R}^{H \times W}$ are regression score map labels.

To this end, the model predictor generates the refined current frame features $\mathit{z_{cur}} \in \mathbb{R}^{H\times W\times C}$ 
and the tracking model $\omega \in \mathbb{R}^{1\times C}$, 
where 
$C$ is the number of channels.
The ClsDec takes $\mathit{z_{cur}}$ and $\omega$ as inputs, convolving them to output a predicted classification score map $ p\in\mathbb {R}^{H\times W}$:
\begin{equation}
\mathit{p} = \omega * \mathit{z_{cur}}.
\label{eqn:basic_score_map}
\end{equation}
\noindent Then, a predicted regression map is computed using RegDec:
\begin{equation}
\mathit{d} = \text{RegDec}\left((\omega \ast \mathit{z_{cur}}) \cdot \mathit{z_{cur}}\right)
\label{eqn:basic_reg_map},
\end{equation}
where operator $\cdot$ denotes channel-wise broadcasting multiplication, and RegDec uses four separate convolution layers to predict four feature maps \(\mathit{d} \in \mathbb{R}^{H\times W\times 4}\) in the \emph{ltrb} 
bounding box representation~\cite{FCOS}.
The coordinates with the highest score in \(\mathit{p}\) are mapped to the regression score map \(\mathit{d}\) for box coordinate prediction.
These processes are used in~\cite{PiVOT,ToMP}.

\subsection{GOT-JEPA for Model Predictor Pre-training}
\label{sec:GOT-JEPA}
In the tracking-by-detection paradigm, an accurate tracking model prediction is crucial for successful tracking. 
Our GOT-JEPA aims to maximize the model prediction capability at this stage by leveraging the JEPA framework~\cite{JEPA} with objectives specialized for tracking.
As shown in~\cref{fig:main_figure}(a), our framework features a teacher network (t-Predictor) and a student network (s-Predictor).
The s-Predictor incorporates a linear network
(ProjNet) tail that learns to predict tracking models capable of accounting for frame variation, as mentioned in \cref{sec:intro}.

To learn a robust model predictor, we initialize the teacher predictor (t-Predictor) from a pretrained tracking model predictor~\cite{ToMP,PiVOT} and keep it fully frozen during GOT-JEPA pre-training to generate pseudo-tracking models $\hat{\omega}$. We freeze the teacher because our objective uses teacher predictions from a clean current frame as pseudo supervision for a student trained on a corrupted current frame under the same history. A frozen teacher keeps pseudo targets stable and prevents the targets from drifting as the student changes, which reduces collapse risk from teacher--student co-adaptation. This principle is supported by recent frozen-teacher designs in self-distillation and representation learning/distillation~\cite{Rethinking_JEPA,Born_Again,DINOv2,WeCoLoRA}.

The student predictor (s-Predictor) is trained to predict $\hat{\omega}$ from corrupted inputs. Both predictors share identical historical conditioning (\ie reference frames and reference labels), but differ in the current-frame observation: the teacher uses a clean frame, while the student uses its corrupted counterpart. This information asymmetry forces the student to match the same pseudo-tracking model under occlusions and distractors, thereby learning invariance and relying on stable target evidence. Since corruptions/augmentations are applied only to the student branch, the model is repeatedly trained on diverse corrupted views that correspond to the same teacher pseudo model, providing diverse supervision without updating the teacher. The lightweight projector in the student branch helps handle corruption-related variation and improves alignment to $\hat{\omega}$, strengthening robustness under adverse observations.
The lightweight projector, termed ProjNet, is a linear network appended to the s-Predictor that generates the adaptive tracking model \(\omega\).
Functioning like a hypernetwork~\cite{Hypernetworks,meta_Hypernetworks,Attentio_hypernetwork}, this predictor dynamically generates the weights used by the localization decoders (which we simplify as prediction tracking models in this paper).
The student learns by aligning its generated model \(\omega\) with the teacher's model \(\hat{\omega}\) using the invariant loss \(\mathcal{L}_{\text{inv}}(\cdot)\), which is computed over $n$ batches:
\begin{equation} \label{eq:invariance}
\mathcal{L}_{\text{inv}}(\omega, \hat{\omega}) = \frac{1}{n} \sum_{i=1}^{n} \| \omega_{i} - \hat{\omega}_{i} \|_{2}^{2}.
\end{equation}

Meanwhile, an additional tail, the Expander \( \mathit{Exp}(\cdot) \), is appended to s-Predictor.
It consists of a $1 \times 1$ convolution layer after s-Predictor and is used to expand the output channel while further optimizing the predictive capability of the tracking models with a covariance loss $\mathcal{L}_{\text{cov}}$~\cite{VICReg} as follows:
\begin{equation} \label{eq:cov_loss}
    \mathcal{L}_{\text{cov}}(\omega_{\text{exp}}) = \frac{1}{c} \sum_{i \neq j} \left[ \text{covM}(\omega_{\text{exp}}) \right]_{i,j}^2,
\end{equation}
\noindent where $\text{covM}(\omega_{\text{exp}}) = \frac{1}{n - 1} \sum_{i=1}^{n} (\omega_{\text{exp}, i} - \bar{\omega}_{\text{exp}})(\omega_{\text{exp}, i} - \bar{\omega}_{\text{exp}})^{T}$ is the covariance matrix, $\omega_{\text{exp}}$ is the output of the Expander, \(i\) and \(j\) are the coordinates of the covariance matrix, $\bar{\omega}_{\text{exp}} = \frac{1}{n} \sum_{i=1}^{n} \omega_{\text{exp}, i}$, $\omega_{\text{exp}} = \mathit{Exp}(\omega)$, and $n$ represents the number of tracking models, each with a dimension of $c$ for the input batches. 
This loss encourages the off-diagonal coefficients of \( \text{covM}(\cdot) \) to approach zero, thereby reducing redundant information in the predicted tracking model. With large-scale, diverse tracking data, redundancy reduction encourages the model to learn more diverse and discriminative patterns, thereby improving the robustness and generalization of model predictions during GOT-JEPA pre-training.
While prior works~\cite{VICReg, OPL} apply similar losses to representation learning, we utilize them for tracking-model prediction to enhance robustness to frame variation across arbitrary targets.

The overall objective function for GOT-JEPA, used in the tracking model prediction learning, is a weighted sum of the invariance and covariance terms:
\begin{equation} \label{eq:got_jepa_loss}
    \mathcal{L}_{\text{mp}} = \alpha \, 
    \mathcal{L}_{\text{inv}}(\omega, \hat{\omega}) + 
    \beta \, \mathcal{L}_{\text{cov}}(\omega_{\text{exp}}),
\end{equation}
where $\alpha$ and $\beta$ are hyperparameters controlling the importance of loss terms. Hyperparameter details are given in~\cref{sec:exp}.

The model predictor enhances tracking model predictions by learning diverse prediction patterns from GOT-JEPA pre-trained, extending beyond simple self-distillation~\cite{SDILS,Born_Again,DINO,schmarje2022one,WeCoLoRA}.
Once the student model predictor is trained,  
as illustrated in~\cref{fig:main_figure}(b),  
the classification and regression decoders (ClsDec and RegDec) are jointly fine-tuned  
for object localisation learning, where this stage validates whether the JEPA-trained model predictor benefits GOT.

\vspace{-0.1 in}

\subsection{OccuSolver}
\label{sec:OccuSolver}
As depicted in~\cref{fig:main_figure}(c), OccuSolver aims to enhance the GOT tracker by providing additional occlusion perception. 
This capability empowers our GOT-JEPA to further achieve superior tracking performance through pixel-level visibility information from the OccuSolver, consequently yielding higher-quality pseudo-reference labels for subsequent frames. 
It enables the model predictor of our GOT-JEPA to incrementally generate more accurate and refined tracking models.

OccuSolver uses object priors (\ie the reference labels) from the GOT tracker and an image sequence as its input. It then refines these priors with the Point Tracker~\cite{Cotracker} 
(\ie the frozen components depicted in OccuSolver of~\cref{fig:main_figure}(c)), and subsequently generates a more accurate output from the Point Tracker that can benefit GOT.

\noindent 
\textbf{Refining the point tracker to be object-centric.}
Given the point-centric nature and lack of object awareness in point trackers, where the initial query points are randomly sampled within the bounding box of the target object in the first frame of the sequence, adaptation is crucial for their utilization in GOT.
To this end, we condition the Point Tracker with two object priors ($p_a$ and $p_b$), identical to the reference labels used for the GOT trackers~\cite{ToMP,PiVOT}.
The object priors are processed by a \emph{Prior Encoder}
(like the label encoder used in ToMP~\cite{ToMP}), which encodes the object priors'
features.
These features are added element-wise to the image features of the first and middle frames in the Point Tracker sequence, respectively.

Specifically, for each query point in a frame sequence, OccuSolver first uses the image encoder of a pretrained point tracker (\ie Cotracker\cite{Cotracker}) to compute the appearance features for each point on each frame.
For query points across time, these features are concatenated spatially to form an appearance token \(Q \in \mathbb{R}^{F}\).
The appearance token is concatenated with the point track (\ie the points' coordinates), $PT \in \mathbb{R}^{2}$.
The concatenated token serves as the input to an iterative transformer (\(\mathit{iter\text{-}Trans}\)), which acts as a non-trainable component positioned above the \emph{Point Head}, and refines the query point coordinates \(PT\) and appearance features \(Q\):
\begin{equation} \label{eq:iter_Trans}
O(PT^{(m+1)}, Q^{(m+1)}) = \textit{iter-Trans}(PT^{(m)}, Q^{(m)}),
\end{equation}
where $m = 1,2,...,M$ indexes the iteration, $PT^{(m)}$ and $Q^{(m)}$ are the refined point track and appearance features, respectively, and $O$ represents the output tokens from the iterative Transformer.
After iterations, the refined point track at the last iteration $\Delta PT$ is fed into a non-trainable \emph{Point Head} for coordinate generation, while the refined appearance features $\Delta Q$ are passed to a learnable VisHead for visibility state estimation. Both the Point Head and VisHead, appended after the iterative Transformer, are adopted from CoTracker~\cite{Cotracker}.

To enable the Point Tracker to utilize the object priors and benefit GOT, we pass each iterative points' appearance features through a four-head two-layer transformer ($\mathit{light\text{-}Trans}$). This network is fine-tuned using the Ladder Side Network~\cite{LST}, which is known for its efficient tuning of large networks and serves as the trainable component positioned above the \emph{Point Head}.
For each query point, the process is formulated as:
\begin{equation} \label{eq:light_Trans}
{\hat Q^{(m)}} = \mathit{light\text{-}Trans}(Q^{(m)}),
\end{equation}
where $\hat Q^{(m)}$ is the output from the $m$-th iteration pass of $\mathit{light\text{-}Trans}$ (like the architecture used in~\cite{DETR}, but with a reduced number of transformer heads and layers).
For memory efficiency, outputs from each $\mathit{light\text{-}Trans}$ layer undergo dimension reduction before tuning.
After this iterative process, the outputs are scaled by $\mathit{ScaleNet}$, a convolution network akin to MLP-Mixer's Mixer Layer~\cite{MLP-Mixer}, and conditioned on the iterative transformer's final output. The refined output $Q_{\text{cond}}$ is then computed:
\begin{equation} \label{eq:ScaleNet}
Q_{\text{cond}} = \hat{Q} + \Delta Q,\; \mbox{where} \; \hat{Q} = \mathit{ScaleNet}\left(\sum_{m=1}^M  \hat{Q}^{(m)}\right).
\end{equation}
Finally, the refined $Q_{\text{cond}}$ acts as the input to \emph{VisHead} for visibility estimation. Together with the Point Head, OccuSolver predicts points' coordinates and their visibility statuses.

\noindent \textbf{Connecting OccuSolver with GOT.}
While OccuSolver enhances the Point Tracker's ability to track objects, the underlying point tracking method faces two key limitations. First, it is prone to failure when objects move dramatically between frames, as each point is tracked within a constrained local search region. Second, for computational efficiency, we track a sparse set of query points (\eg 128 points), which can yield a coarse representation that is not fine-grained across the object's entire boundary.
To mitigate these issues, we introduce an Ensemble Network, a four-head two-layer transformer architecture, that learns to modulate the sparse point visibility features from OccuSolver with the dense visual features of the current frame from the GOT, creating a more robust and complete target representation.

We first introduce a \emph{Mapping Function} 
to transform discrete OccuSolver point coordinates into a dense spatial representation. 
For each of the $C$ predicted point coordinates, a Gaussian kernel is applied to generate an initial energy map $e$.
Details of the Gaussian function generation can be found in~\cite{danelljan2016beyond}.
If a point is invisible, its map is negated ($1 - e$). 
These individual energy maps are concatenated to form $\mathbf{E}\in \mathbb{R}^{H \times W \times C}$, which is then spatially concatenated with the GOT current frame feature $z_{\text{cur}}$ and processed by a lightweight transformer (\ie the aforementioned Ensemble Network).
This step models the interplay between visual and visibility cues, yielding a visibility-aware current frame features $\tilde{\mathbf{E}} \in \mathbb{R}^{H \times W \times C}$.

The \emph{Ensemble Network}, \(\mathcal{E}(\cdot, \cdot)\), then modulates between the original current frame feature \(\mathbf{z}_{\text{cur}}\) and the visibility-aware feature \(\tilde{\mathbf{E}}\).
Here, \(\mathbf{z}_{\text{cur}}\) denotes the current frame feature generated by the \emph{Model Predictor}, as shown in the figure.
This produces a refined feature
$\tilde{z}_{\text{cur}} = \mathcal{E}(\tilde{\mathbf{E}}, z_{\text{cur}})$.
The training objective for OccuSolver is defined as follows:

\begin{equation}
\begin{aligned}
\mathcal{L} = &\; \lambda_{\text{cpt}}
\mathcal{L}_{\text{cls}}
(\hat{p}, p_{\text{pt}}) +
\lambda_{\text{rpt}}
\mathcal{L}_{\text{reg}}
(\hat{d}, d_{\text{pt}}) \\
& + \lambda_{\text{cgot}}
\mathcal{L}_{\text{cls}}
(\hat{p}, p_{\text{got}}) +
\lambda_{\text{rgot}}
\mathcal{L}_{\text{reg}}
(\hat{d}, d_{\text{got}}),
\label{eqn:total_GOT-JEPA_PT_loss}
\end{aligned}
\end{equation}
where the total loss is weighted by $\lambda_{\text{cpt}}$, $\lambda_{\text{rpt}}$, $\lambda_{\text{cgot}}$, and $\lambda_{\text{rgot}}$.

Similar to~\cref{eqn:basic_score_map} where the score map $p$ is obtained by $\mathit{p} = \omega * \mathit{z_{\text{cur}}}$, $\mathit{z_{\text{cur}}}$ is replaced by $\tilde{\mathbf{E}}$ and $\tilde{\mathbf{z}}_{\text{cur}}$ in~\cref{eqn:total_GOT-JEPA_PT_loss} to obtain $\mathit{p_{\text{pt}}}$ and $\mathit{p_{\text{got}}}$, respectively.
Similarly, in~\cref{eqn:basic_reg_map}, the regression map $d$ is given by $\mathit{d} = \text{RegDec}\left((\omega \ast \mathit{z_{\text{cur}}}) \cdot \mathit{z_{\text{cur}}}\right)$. Here, $\mathit{z_{\text{cur}}}$ is substituted by $\tilde{\mathbf{E}}$ and $\tilde{z}_{\text{cur}}$ to generate $\mathit{d_{\text{pt}}}$ and $\mathit{d_{\text{got}}}$.
The tracking model $\omega$ for the above process is generated by the model predictor. This dual-supervision strategy ensures that the visibility features learned from OccuSolver are effectively integrated into the generic object tracker, enhancing the tracker's robustness to occlusion and deformation.


\begin{table*}
\centering
\caption{Comparison of our method with SOTA trackers on various datasets. All trackers utilise the same training data for consistency. ``*'' indicates adherence to specific GOT-10k (GOT) guidelines~\cite{GOT-10k}. For existing methods, we report their highest-resolution variant, prioritising ViT-L backbones when available, otherwise their best-performing variant.
}
\resizebox{0.82\textwidth}{!}{%
\begin{tabular}{lc|cccc|ccccc}
\hline
\multicolumn{2}{c|}{Training-Test Class Overlap} & \multicolumn{4}{c|}{Low or No Overlap} & \multicolumn{5}{c}{Full Overlap} \\ \hline
\multicolumn{2}{c|}{Dataset} & AVisT & NfS & OTB & GOT* & \multicolumn{3}{c}{LaSOT} & \multicolumn{2}{c}{TrackingNet} \\ \hline
\multicolumn{1}{c|}{Tracker} & Year & SUC & SUC & SUC & AO & NPr & Pr & SUC & NPr & SUC \\ \hline
\multicolumn{1}{l|}{\textbf{GOT-JEPA (Ours)}} & - & \textbf{63.7} & \textbf{70.8} & \textbf{73.2} & \textbf{79.6} & \textbf{85.3} & \textbf{83.2} & \textbf{75.4} & \textbf{90.6} & \textbf{86.4} \\
\multicolumn{1}{l|}{UniSOT~\cite{UniSOT}} & 2026 & 57.8 & 67.6 & - & - & - & 78.3 & 71.3 & - & 84.1 \\
\multicolumn{1}{l|}{SAMURAI~\cite{SAMURAI}} & 2026 & - & 69.2 & 71.5 & 81.7 & 82.7 & 80.2 & 74.2 & - & 85.3 \\
\multicolumn{1}{l|}{PiVOT~\cite{PiVOT}} & 2025 & 62.2 & 68.2 & 71.2 & 76.9 & 84.7 & 82.1 & 73.4 & 90.0 & 85.3 \\
\multicolumn{1}{l|}{CVT-Track~\cite{CVT-Track}} & 2025 & - & - & - & 74.9 & 80.4 & 77.9 & 71.2 & 88.2 & 84.2 \\
\multicolumn{1}{l|}{MPIT~\cite{MPIT}} & 2025 & - & 66.9 & 70.9 & 73.3 & 78.2 & 73.7 & 69.4 & 88.1 & 83.3 \\
\multicolumn{1}{l|}{USCLTrack~\cite{USCLTrack}} & 2025 & - & - & 70.8 & 75.8 & 82.4 & 89.9 & 73.5 & 89.4 & 84.9 \\
\multicolumn{1}{l|}{MFDSTrack~\cite{MFDSTrack}} & 2025 & - & 66.8 & 66.8 & 73.6 & 81.2 & 77.6 & 72.1 & 88.5 & 84.2 \\
\multicolumn{1}{l|}{MGTrack~\cite{MGTrack}} & 2025 & - & - & - & 76.2 & 81.0 & 84.6 & 84.7 & 89.7 & 84.7 \\
\multicolumn{1}{l|}{SATrack~\cite{SATrack}} & 2025 & 58.4 & 67.5 & - & 75.4 & 81.4 & 78.4 & 72.0 & 89.0 & 84.7 \\
\multicolumn{1}{l|}{SuperSBT~\cite{SuperSBT}} & 2024 & - & 67.7 & 68.9 & 75.5 & 82.5 & 78.6 & 72.8 & 88.9 & 84.8 \\
\multicolumn{1}{l|}{LoRAT~\cite{LoRAT}} & 2024 & 62.0 & 66.7 & 72.0 & 77.5 & 84.1 & 82.0 & 75.1 & 89.7 & 85.6 \\
\multicolumn{1}{l|}{ARTrackv2~\cite{ARTrackV2}} & 2024 & - & 68.4 & - & 79.5  & 82.8 & 81.1 & 73.6 & 90.4 & 86.1 \\
\multicolumn{1}{l|}{HIPTrack~\cite{HIPTrack}} & 2024 & - & 68.2 & 71.0 & 74.5 & 82.9 & 79.5 & 72.7 & 89.1 & 84.5 \\
\multicolumn{1}{l|}{SiamON~\cite{SiamON}} & 2023 & - & - & 64.4 & - & - & - & - & - & - \\
\multicolumn{1}{l|}{ROMTrack~\cite{ROMTrack}} & 2023 & 59.1 & 67.5 & 71.4 & 74.2 & 81.4 & 78.2 & 71.4 & 89.0 & 84.1 \\
\multicolumn{1}{l|}{GRM~\cite{GRM}} & 2023 & 54.5 & 66.9 & 68.9 & 73.4 & 81.2 & 77.9 & 71.4 & 88.9 & 84.0 \\
\multicolumn{1}{l|}{OSTrack~\cite{OSTrack}} & 2022 & 57.7 & 66.5 & 68.1 & 73.7 & 81.1 & 77.6 & 71.1 & 88.5 & 83.9 \\ 
\multicolumn{1}{l|}{ToMP~\cite{ToMP}} & 2022 & 50.9 & 67.0 & 70.1 & - & 79.2 & 73.5 & 68.5 & 86.4 & 81.5 \\
\hline
\end{tabular}
}
\label{tab:GOT-JEPA-SOTAs}
\end{table*}

\section{Experiments}
\label{sec:exp}


%

This section evaluates our method, covering experimental settings, SOTA comparison, and ablation studies.


\subsection{Experimental Setting}

\noindent \textbf{Training Data.}
Like most trackers, e.g.~\cite{ToMP,PiVOT,SeqTrack,LoRAT}, we adopt the training splits of LaSOT, GOT10k, TrackingNet, and COCO for model training.
The training data rigorously follows the VOT2022 challenge and GOT-10K guidelines.

\noindent \textbf{Test Data.}
We use the following datasets for evaluation:
\begin{itemize}

\item \textbf{AVisT}~\cite{AVisT}: It is designed for testing without a training set, encompassing 120 short and long sequences, averaging 664 frames each under adverse visibility conditions.

\item \textbf{NfS}~\cite{NfS} and \textbf{OTB-100}~\cite{OTB}:
They are used for testing without a corresponding training set, each containing 100 sequences with an average of 534 frames per sequence.

\item \textbf{GOT-10k}~\cite{GOT-10k}: It has $420$ short sequences with an average of 149 frames per sequence, featuring non-overlapping object classes in the training and test sets. 
%

\item \textbf{LaSOT}~\cite{LaSOT} and \textbf{TrackingNet}~\cite{LaSOT}: 
They provide training data where test classes fully overlap with training classes.
LaSOT provides $280$ long sequences with an average of 2k frames per sequence
TrackingNet offers $511$ short sequences, averaging 471 frames each.

\item \textbf{VOT2022}~\cite{VOT2022}:  
It is the 2022 edition of the Visual Object Tracking short-term box (VOT-STb2022) challenge. 
%
\end{itemize}
%
%

\textbf{Evaluation Metrics.}
Like recent methods~\cite{PiVOT, LoRAT}, we evaluate trackers using the following metrics:
\begin{itemize}

\item \textbf{SUC} (success rate): The percentage of frames in which the predicted bounding box overlaps the ground truth by at least an IoU threshold or the average of all thresholds.

%

\item \textbf{Pr} (precision): It measures the percentage of frames where the predicted target center is within $T$ pixels of the ground-truth center. $T$ is set to $20$ in this work.

\item \textbf{NPr} (normalized precision): 
It is the percentage of frames where the center location error normalized by the target’s box diagonal is less than a threshold $0.2$.

\item \textbf{AO} (average overlap): It represents the mean IoU between the predicted and ground-truth bounding boxes.

\end{itemize}

\noindent \textbf{Training Procedure.}
Our method involves two main stages: 
1) GOT-JEPA training and 2) OccuSolver training.




At Stage 1, 
we first pre-train the Model Predictor using our JEPA architecture. 
After pre-training, it is fine-tuned with the Tracking Head.
Specifically, during pre-training, 
a student predictor learns to predict tracking models from corrupted inputs. 
A teacher predictor, derived from the ToMP-L (a ViT-L variant of ToMP~\cite{ToMP,PiVOT}) model predictor, 
provides pseudo-tracking models from non-corrupted inputs. 
This setup allows the student predictor to learn from superior pseudo-tracking models and adapt to more challenging input cases.

The pre-training stage focuses solely on pre-training the model predictor to generate tracking models, 
without involving the Tracking Head or localisation objectives.
After the Model Predictor is pre-trained, the Tracking Head is trained jointly.
The tracking performance of GOT-JEPA without OccuSolver is then assessed (Table \ref{tab: sALLdataset_tompL_JEPA_Cov_PT}, row (4)).


At stage 2,  
we train OccuSolver with the model predictor and Tracking Head.  
The model predictor and Tracking Head are initialized with weights trained at stage 1, and their model weights are frozen in this stage.
This stage investigates whether OccuSolver improves the tracker compared to the tracker trained at stage 1.
OccuSolver processes a 
$8$ consecutive frames from a sequence.
The Tracking Head treats frames $1$ and $5$ as the reference frames, and frame $8$ as the current frame, along with their corresponding labels.



\begin{table}[!t]
\caption{Comparisons among different trackers on VOT-STb2022.  
SOTA results are reported from the VOT2022 paper~\cite{VOT2022}.
}
\resizebox{0.47\textwidth}{!}{
\begin{tabular}{ccccccc}
\hline
Metrics    & GOT-JEPA       & MixFormerL & OSTrackSTB & TransT\_M & SwinTrack & tomp  \\ \hline
Robustness & 0.898 & 0.859      & 0.869      & 0.849     & 0.803     & 0.818 \\
AUC        & 0.728 & 0.708      & 0.680      & 0.639     & 0.626     & 0.628 \\ \hline
\end{tabular}
}
\vspace{-0.1in}
\label{tab:VOT2022}
\end{table}

\noindent \textbf{Implementation Details.}
Our method is implemented using PyTorch 2.0.0 and CUDA 11.7, and the tracker is built upon the ToMP framework~\cite{ToMP}.
The tracker operates on an NVIDIA RTX 4090 GPU, utilizing approximately 3 GB of GPU memory during evaluation and achieving 24 FPS and 50 FPS for the high- and low-resolution variants, respectively.
In the first training stage, 8 GPUs are employed to train the proposed tracker with the L-378 variant. In the second stage, 4 GPUs are used.
For the L-252 variant, all training stages are conducted using 4 GPUs.
The batch size ranges from 48 to 64, maximized according to available computational resources.
In~\cref{eq:got_jepa_loss}, \(\alpha\) is $25$ times higher than \(\beta\).
In~\cref{eqn:total_GOT-JEPA_PT_loss}, we set \(\lambda_{\text{cgot}}=200\), \(\lambda_{\text{cpt}}=100\), \(\lambda_{\text{rgot}}=1\), and \(\lambda_{\text{rpt}}=0.5\), respectively.
We use Copy-Paste~\cite{Copy_Paste,trial} to simulate frame variants such as occlusions and distractor elements. The copy-paste operation is applied in feature space on the patch-feature grid of the current frame. Concretely, after the backbone produces a feature map $F$ with shape $B\times C\times H\times W$, we sample a corruption ratio $\rho \sim \mathcal{U}(0,\rho_{\max})$ with $\rho_{\max}=0.2$ and set $K=\lfloor \rho HW \rfloor$. For each training sample, we randomly select $K$ patch positions on the $H\times W$ grid as the source positions to copy, and randomly choose another $K$ patch positions on the same grid as the target positions to paste. This ensures that the copied-and-pasted regions vary across samples and iterations. The features at the target positions are replaced with those copied from the source positions. This corruption is applied only to the student branch for the current-frame features, while the teacher branch uses clean current-frame features.
AdamW~\cite{ADAMW} is used as the optimization solver.
The learning rate is set to \(10^{-4}\) for most components, while the Expander of the s-Predictor uses a rate of \(10^{-3}\).

\noindent The objective function for target classification is the compound hinge loss of DiMP~\cite{DiMP}, while the GIoU loss~\cite{GIOU} is used for target regression.
In line with recent works such as PiVOT~\cite{PiVOT} and LoRAT~\cite{LoRAT}, we employ ViT-L as the backbone for image feature extraction, utilizing weights pretrained with DINOv2~\cite{DINOv2}. The backbone remains frozen during tracker training.
These settings are used in all experiments.
Further details are in the supplementary material.
To mitigate the computational cost of using higher image resolutions, as is common practice in recent works~\cite{TemTrack,MambaLCT,SeqTrack,LoRAT}, we use lower resolutions for ablation studies but employ higher resolutions for comparison with SOTAs:
1) \textbf{GOT-JEPA-252}, where the frame resolution and the patch token size are respectively set to $252\times252$ and $18\times18$, and 2) \textbf{GOT-JEPA-378}, where the frame resolution and the token size are respectively set to $378\times378$ and $27\times27$.


\begin{figure*}[!t]
	\centering
 
        \begin{tabular}{ccc}

        
        {{\includegraphics[scale=0.28]
        {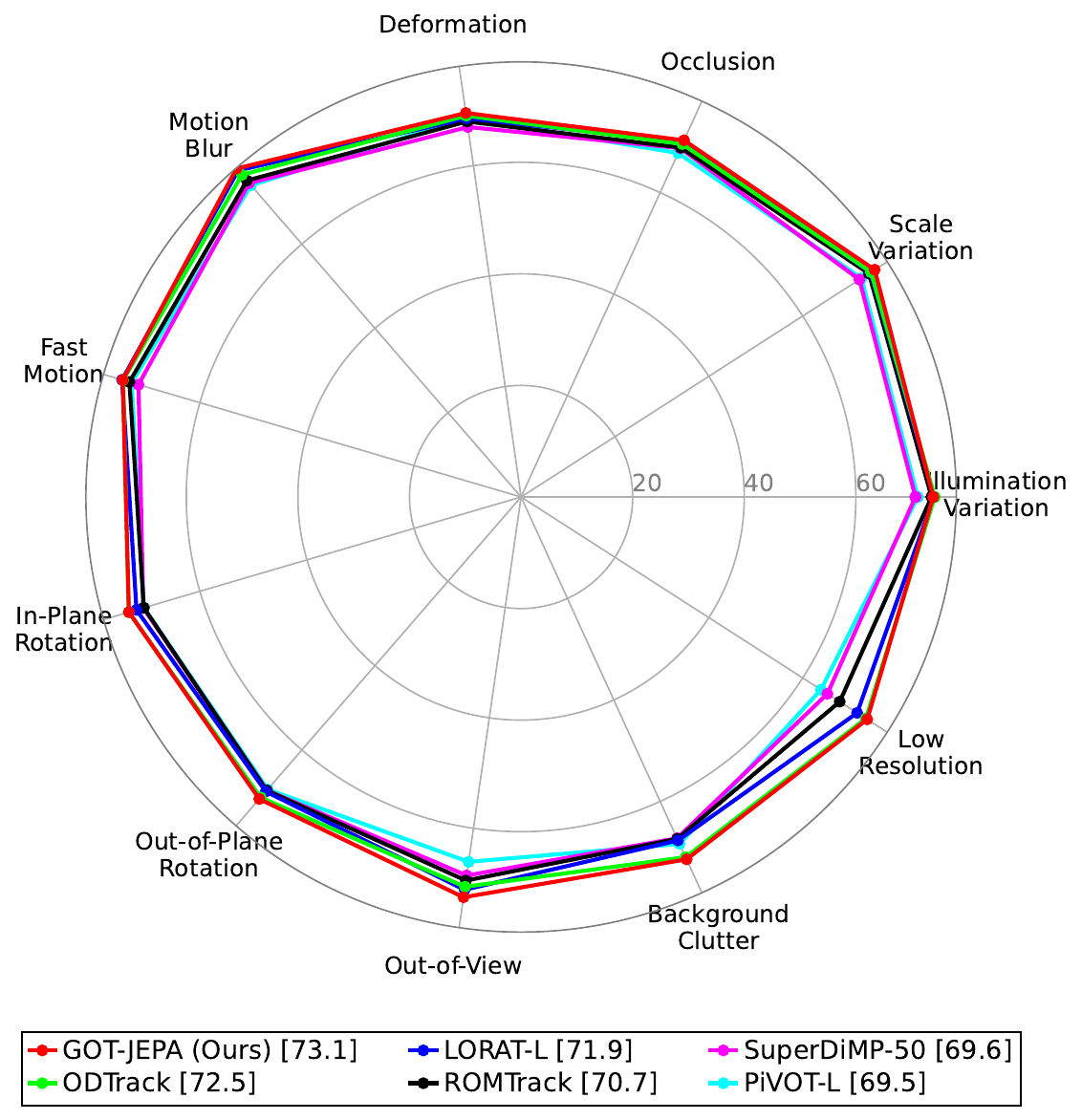}}}&

        {{\includegraphics[scale=0.28]
        {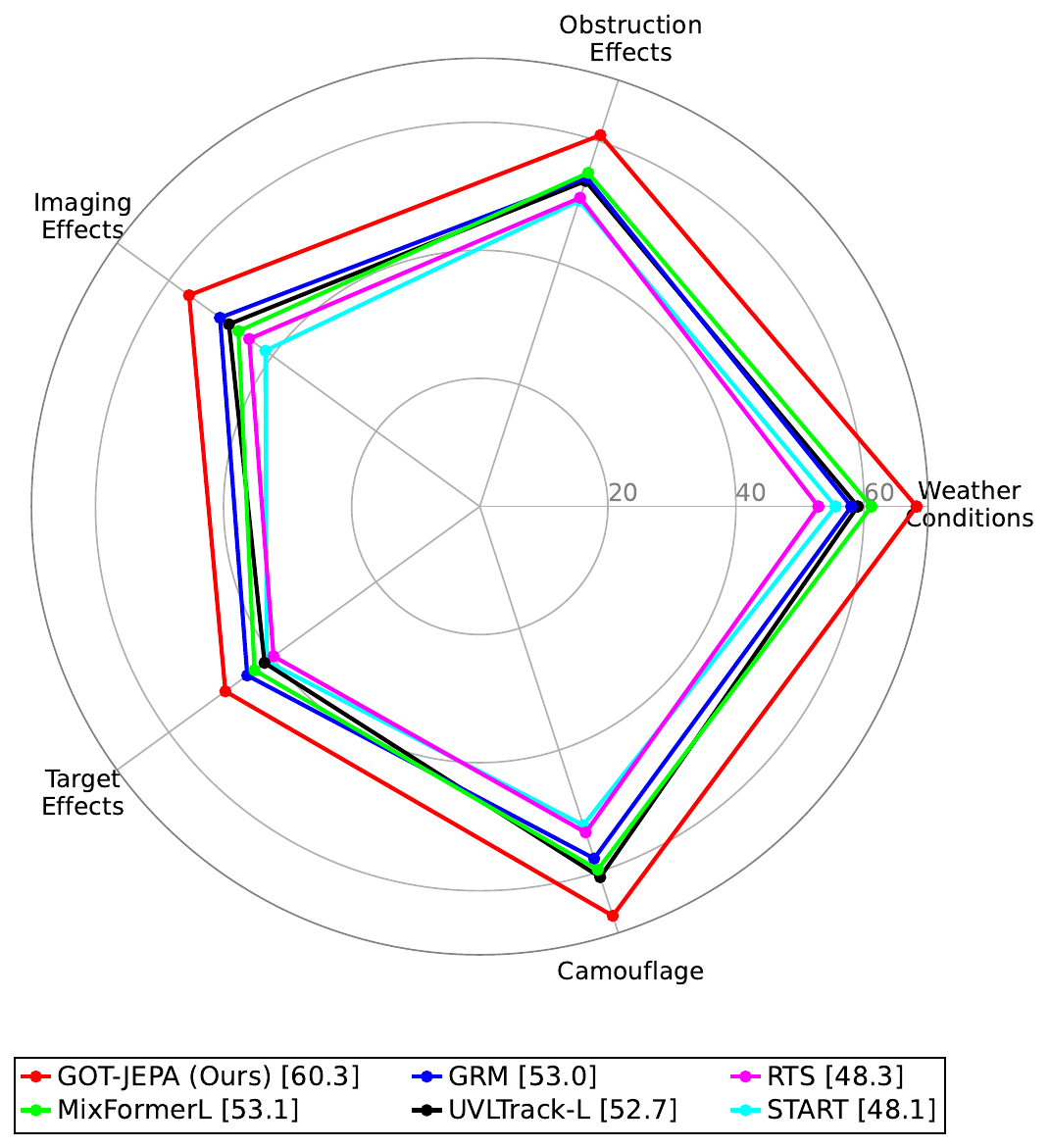}}}&

        {{\includegraphics[scale=0.28]
        {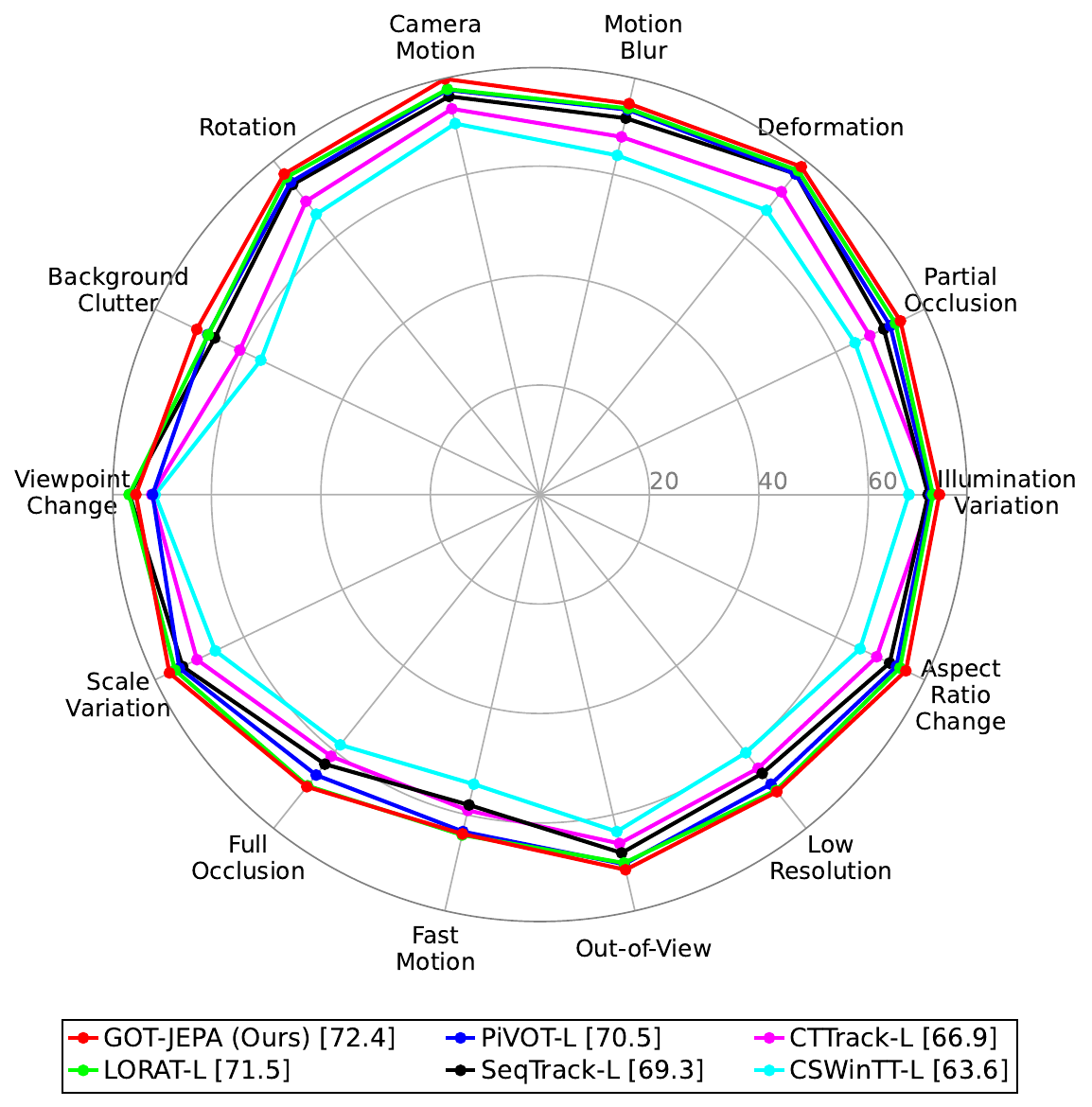}}}
        
        \end{tabular}
	\caption{This figure presents the attribute analysis of OTB-100, AVisT, and LaSOT from left to right, with the average scores at the bottom.
        }
	\label{fig:radar_plot}
\end{figure*}


\begin{table}[t]
\centering
\color{black}
\caption{Attribute-wise results on OTB, AVisT, and LaSOT using large-resolution variants for several important attributes.}
\label{tab:attribute_results_large_res}
\scriptsize
\setlength{\tabcolsep}{3pt}
\renewcommand{\arraystretch}{1.08}

\resizebox{0.47\textwidth}{!}{%
{
\begin{tabular}{llccccc}
\toprule
\multirow{2}{*}{Dataset} & \multirow{2}{*}{Tracker} &
\thead{Background\\Clutter} &
\thead{Deformation} &
\thead{Occlusion} &
\thead{Out-of-View} &
\thead{Target\\Visibility} \\
\midrule

\multirow{4}{*}{OTB}
& \textbf{Ours}   & 71.4 & 69.5 & 70.3 & 72.5 & -- \\
& ToMP-L           & 68.2 & 67.1 & 67.3 & 69.4 & -- \\
& PiVOT          & 68.4 & 68.6 & 67.8 & 66.1 & -- \\
& ROMTrack       & 67.4 & 68.0 & 68.9 & 69.5 & -- \\
\midrule

\multirow{4}{*}{AVisT}
& \textbf{Ours}   & 67.2 & 49.0 & 61.3 & --   & 67.8 \\
& ToMP-L            & 65.5 & 45.7 & 57.4 & --   & 65.2 \\
& PiVOT          & 67.1 & 47.7 & 59.2 & --   & 66.5 \\
& MixFormer      & 59.6 & 43.5 & 54.8 & --   & 61.2 \\
\midrule

\multirow{4}{*}{LaSOT}
& \textbf{Ours}   & 69.6 & 76.6 & 70.7 & 70.3 & -- \\
& ToMP-L            & 66.4 & 74.9 & 67.1 & 68.3 & -- \\
& PiVOT          & 67.3 & 74.9 & 68.3 & 69.3 & -- \\
& ROMTrack       & 64.6 & 73.4 & 66.9 & 67.1 & -- \\
\bottomrule
\end{tabular}%
}%
}
\end{table}


\begin{table}[!t]
\centering
\caption{Comparisons among different trackers on multiple datasets using OP50 as a metric.
}
\resizebox{0.35\textwidth}{!}{
\begin{tabular}{l|ccc}
\hline
\multicolumn{1}{c|}{Dataset} & NfS & AVisT & LaSOT \\ \hline
\multicolumn{1}{c|}{Tracker / Metric} & \multicolumn{3}{c}{OP50} \\ \hline
\textbf{Ours} & 89.60 & 73.67 & 86.46 \\
ToMP-L & 85.69 & 72.55 & 84.78 \\
LoRAT-L & 85.56 & - & 85.11 \\
UVLTrack-L & 85.58 & 65.11 & 83.14 \\
GRM-L & 83.53 & 63.83 & 82.75 \\
SeqTrack-L & 82.37 & - & 82.98 \\ 
\hline
\end{tabular}
}
\label{tab:GOT_JEPA_OP50}
\end{table}

\subsection{Comparisons with the State-of-the-Art Methods}
\cref{tab:GOT-JEPA-SOTAs} compares GOT-JEPA with SOTAs on several benchmark datasets.
`ToMP-L' is the baseline method that utilises a high image resolution.
Our tracker is mostly close to PiVOT~\cite{PiVOT}, where both our tracker and PiVOT use the DiNOv2~\cite{DINOv2} ViT-L variant of ToMP~\cite{ToMP} (\ie ToMP-L) as a base tracker.
LoRAT~\cite{LoRAT} also uses the DiNOv2 ViT-L for image feature, though it uses OSTrack~\cite{OSTrack} as a base tracker.
We also compare our tracker against several other ViT-L-based trackers, including DiffusionTrack~\cite{DiffusionTrack}, SeqTrack~\cite{SeqTrack}, GRM~\cite{GRM}, UVLTrack~\cite{UVLTrack}, CTTrack~\cite{CTTrack}, CSWinTT~\cite{CSWinTT}, and ODTrack~\cite{ODTrack}.
Overall, GOT-JEPA exhibits robust performance and generalizes well to out-of-distribution and in-distribution targets.

%

As shown in~\cref{tab:GOT-JEPA-SOTAs}, GOT-JEPA demonstrates strong generalization against recent state-of-the-art trackers on both out-of-distribution and in-distribution benchmarks. 
For out-of-distribution evaluation, GOT-JEPA achieves a success rate of $63.7\%$ on AVisT, outperforming PiVOT ($62.2\%$), LoRAT ($62.0\%$), and UniSOT ($57.8\%$).
On NfS, GOT-JEPA reaches $70.8\%$, exceeding HIPTrack and PiVOT ($68.2\%$ each) and UniSOT ($67.6\%$).
On OTB, GOT-JEPA attains the best success rate of $73.2\%$, surpassing SAMURAI ($71.5\%$), LoRAT ($72.0\%$), ROMTrack ($71.4\%$), and PiVOT ($71.2\%$).

On GOT-10k, GOT-JEPA achieves the highest AO of $79.6\%$, outperforming LoRAT ($77.5\%$), PiVOT ($76.9\%$), and MGTrack ($76.2\%$). However, our tracker lags behind SAMURAI, which uses SAM 2 for segmentation tracking, while our tracker still performs better than the other trackers across the remaining datasets, except on GOT-10K.
These consistent gains under distribution shift validate the effectiveness of OccuSolver, which leverages object-aware visibility cues for explicit occlusion handling. Furthermore, the results highlight the benefit of GOT-JEPA pre-training in enhancing target-identity preservation over long temporal windows, particularly under adverse visibility conditions.

For in-distribution evaluation, GOT-JEPA remains competitive and achieves the best normalized precision on LaSOT ($85.3\%$ NPr), while also attaining strong accuracy in success rate ($75.4\%$). 
On TrackingNet, GOT-JEPA achieves the best results in both metrics, with $90.6\%$ NPr and $86.4\%$ success rate, outperforming strong baselines such as LoRAT ($89.7\%$ NPr and $85.6\%$ success rate) and PiVOT ($90.0\%$ NPr and $85.3\%$ success rate).

The comparison on the VOT2022 challenge, as shown in~\cref{tab:VOT2022}, further demonstrates that the tracker achieves the highest robustness and AUC scores.



\vspace{+0.2in}

\noindent \textbf{Comparison of Trackers Using OP50 as a Metric}: 
In addition to SUC, NPr, and Pr, we compare trackers using OP50, which measures the percentage of frames where the predicted and ground-truth IoU exceed 50\%. 
The results are presented in~\cref{tab:GOT_JEPA_OP50}. 
On the NfS dataset, our tracker surpasses UVLTrack-L by 4.02\%. 
On the AVisT dataset, it outperforms UVLTrack-L by 8.56\%. 
On the LaSOT dataset, it exceeds LoRAT-L by 1.35\%.



\noindent \textbf{NPr, Pr, and SUC Plots and Analysis:}
We report the Normalized Precision (NPr), Precision (Pr), and Success (SUC) plots on four datasets: NfS, AVisT, LaSOT, and OTB-100. 
Other datasets, including VOT2022, TrackingNet, and GOT-10k, are evaluated on online servers without plot outputs and are therefore excluded from this analysis. 
The generation of plots requires raw tracker outputs; hence, methods lacking such data are omitted from the comparison.

\noindent In the Precision (Pr) and Normalized Precision (NPr) plots, the x-axis represents thresholds, while the y-axis denotes the percentage of frames where the predicted target center lies within these thresholds. 
Trackers are typically ranked at 20 pixels in Pr or 0.2 in NPr.  
In the Success (SUC) plot, the x-axis corresponds to IoU thresholds, and the y-axis shows the percentage of frames that exceed these thresholds. 
Trackers are ranked by the average precision across all thresholds.

\begin{figure*}[!t]
	\centering
 
        \begin{tabular}{ccc}

        \hspace{-0.3cm}
        
        {{\includegraphics[scale=0.22]{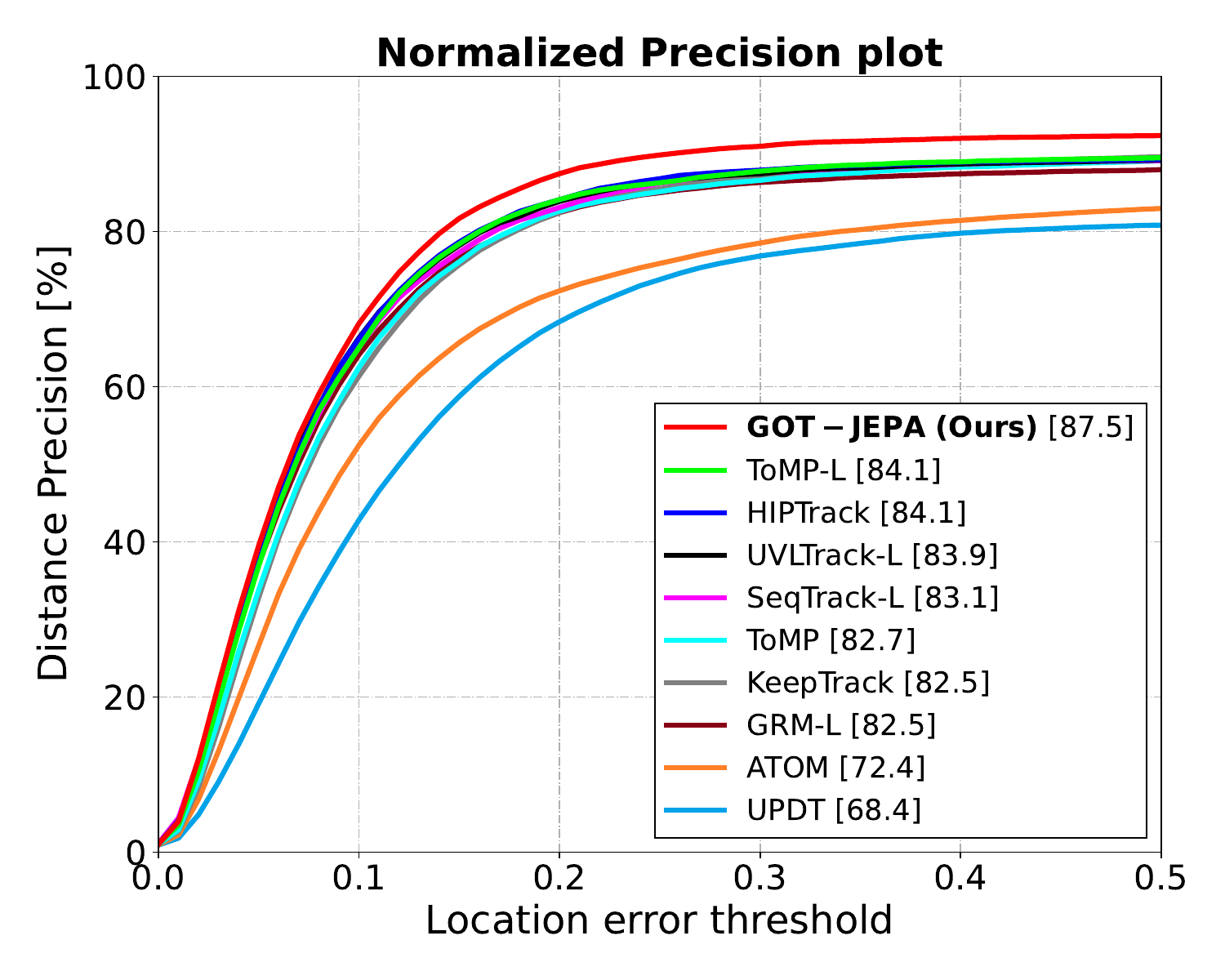}}}&
        
        \hspace{-0.3cm}
        
        {{\includegraphics[scale=0.22]{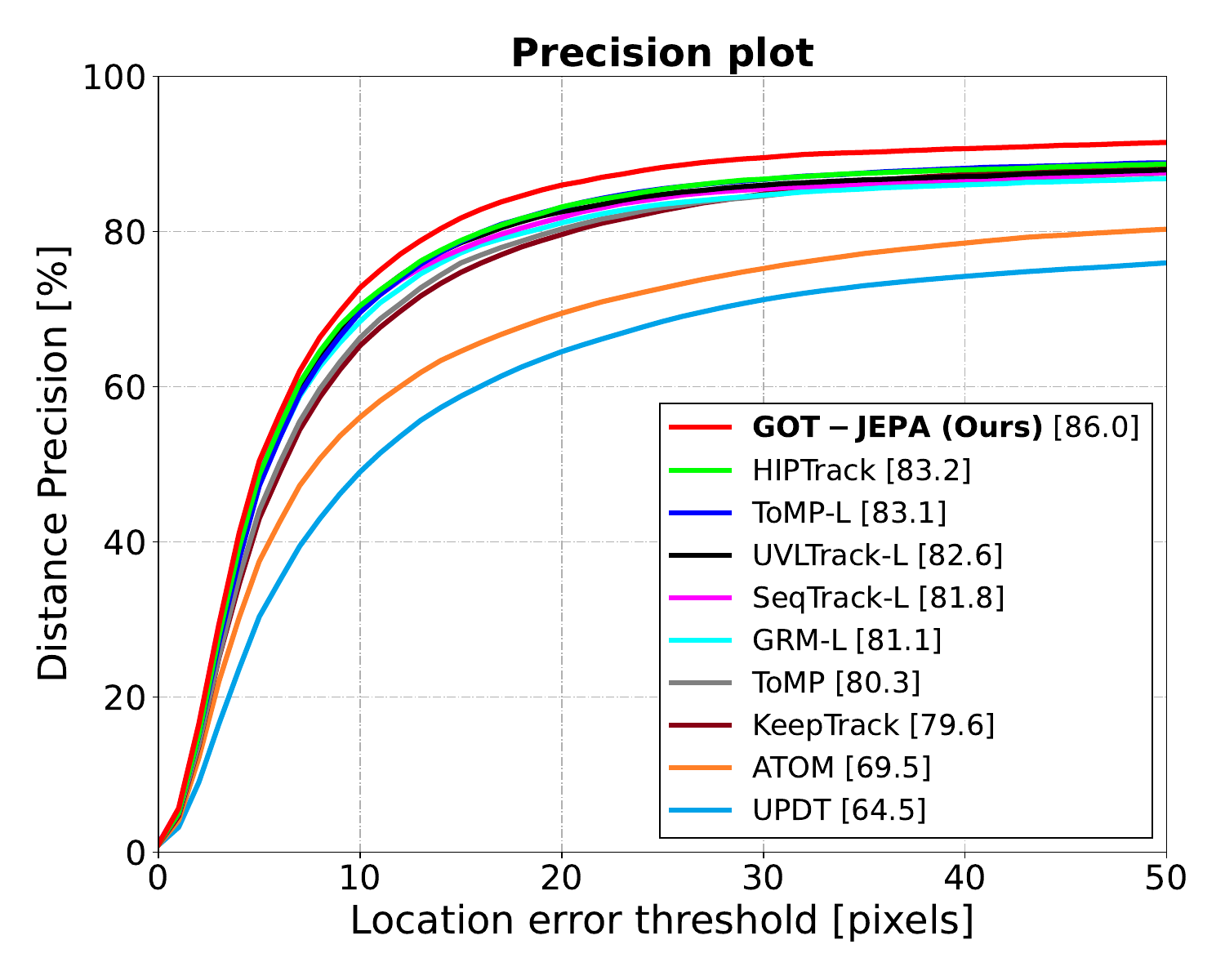}}}&

        \hspace{-0.3cm}
        
        {{\includegraphics[scale=0.22]{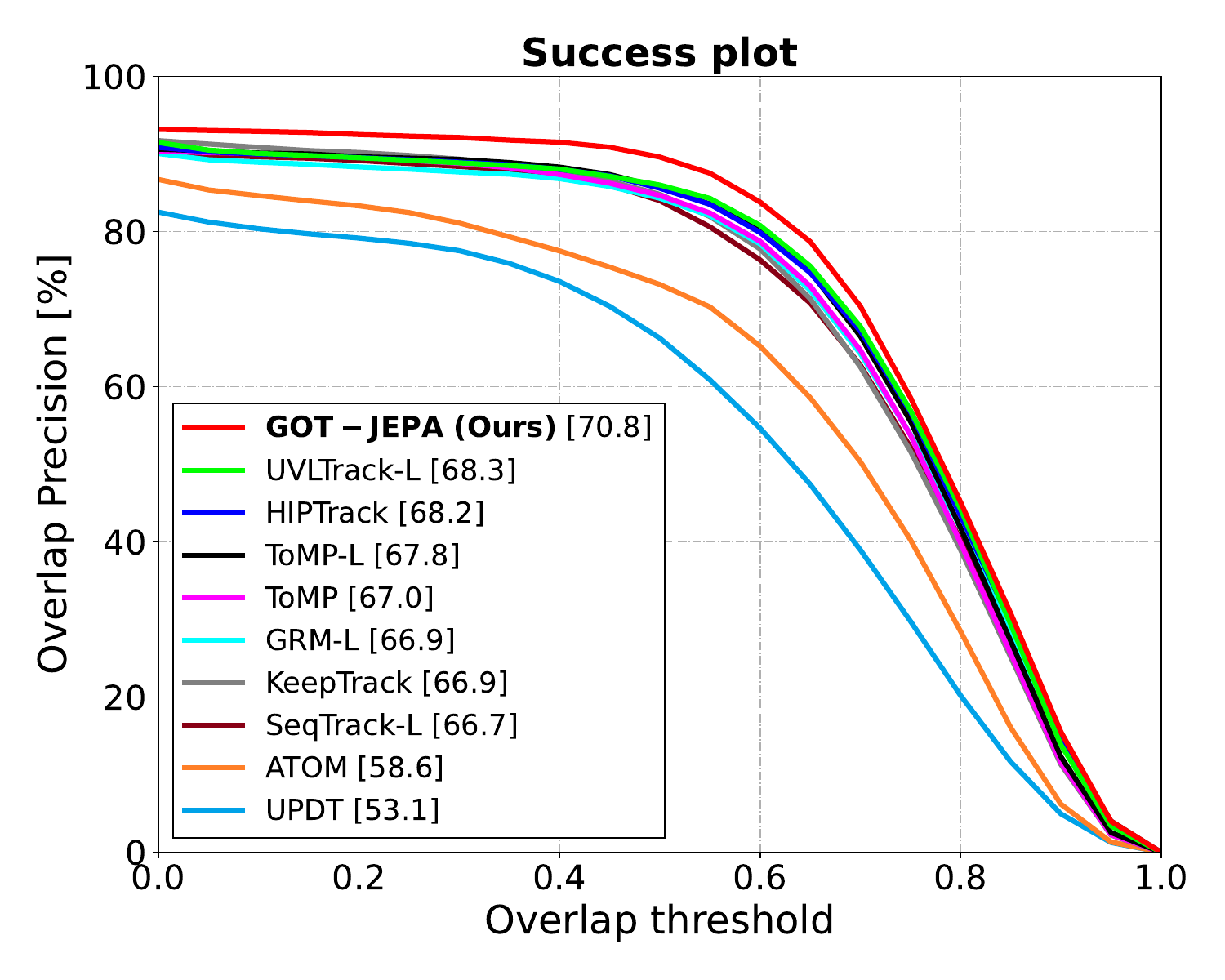}}}
        

        \end{tabular}
        
	\vspace{-0.15cm}
 
	\caption{
        \textbf{
        Comparison of methods using NPr, Pr, and SUC plots on the NfS dataset, from left to right.
        } 
        }
	\label{fig:NfS_all}
\vspace{-0.10in}
\end{figure*}

\noindent The detailed analysis of each dataset is summarized as follows:  


\noindent \textbf{NfS}: As illustrated in~\cref{fig:NfS_all}, our tracker outperforms all competing methods once the threshold exceeds 0.1 in NPr or 10 pixels in Pr. 
For SUC, the proposed method consistently achieves superior performance across all thresholds.  

\begin{figure*}[!t]
	\centering
 
        \begin{tabular}{ccc}

        \hspace{-0.3cm}
        
        {{\includegraphics[scale=0.22]{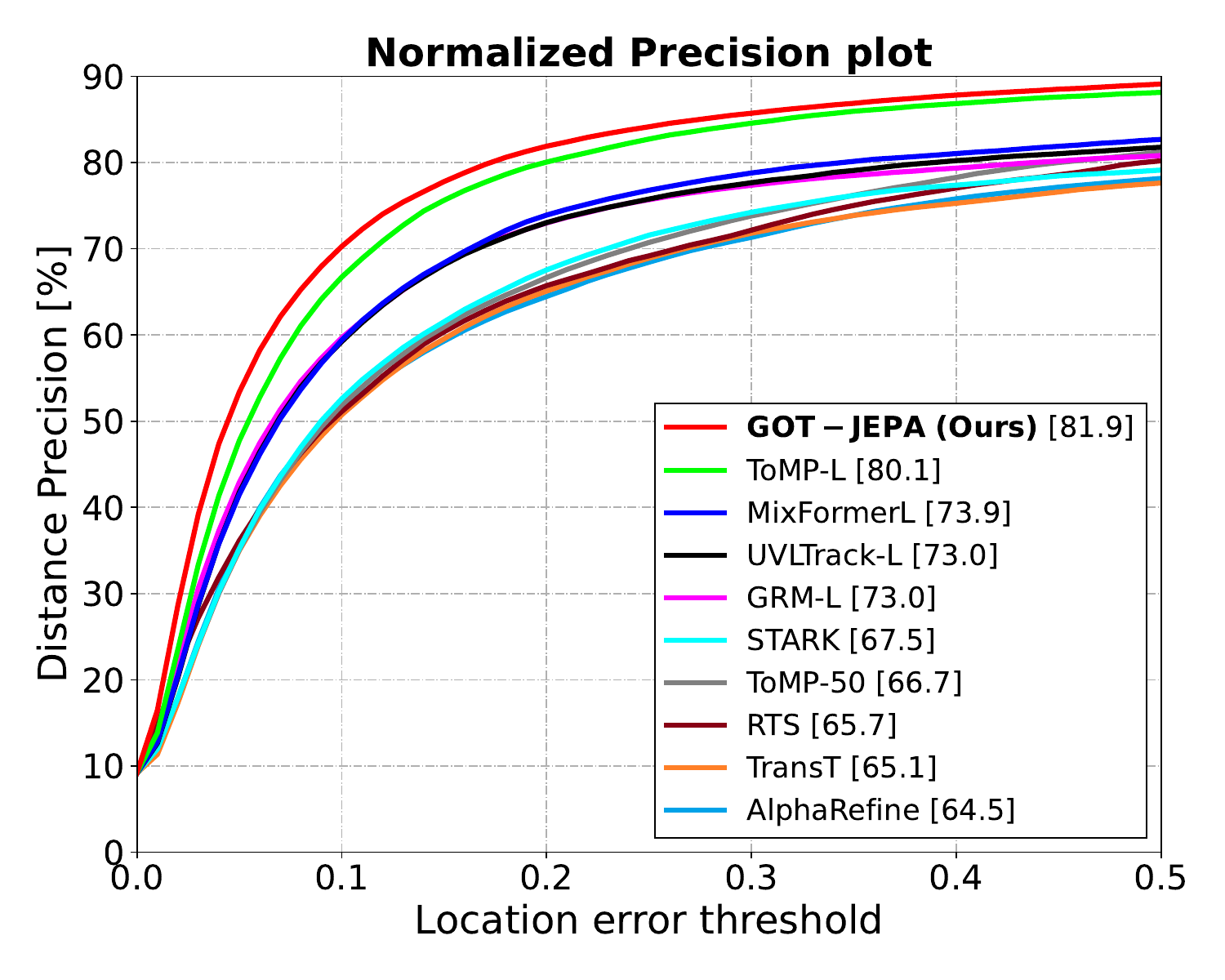}}}&
        
        \hspace{-0.3cm}
        
        {{\includegraphics[scale=0.22]{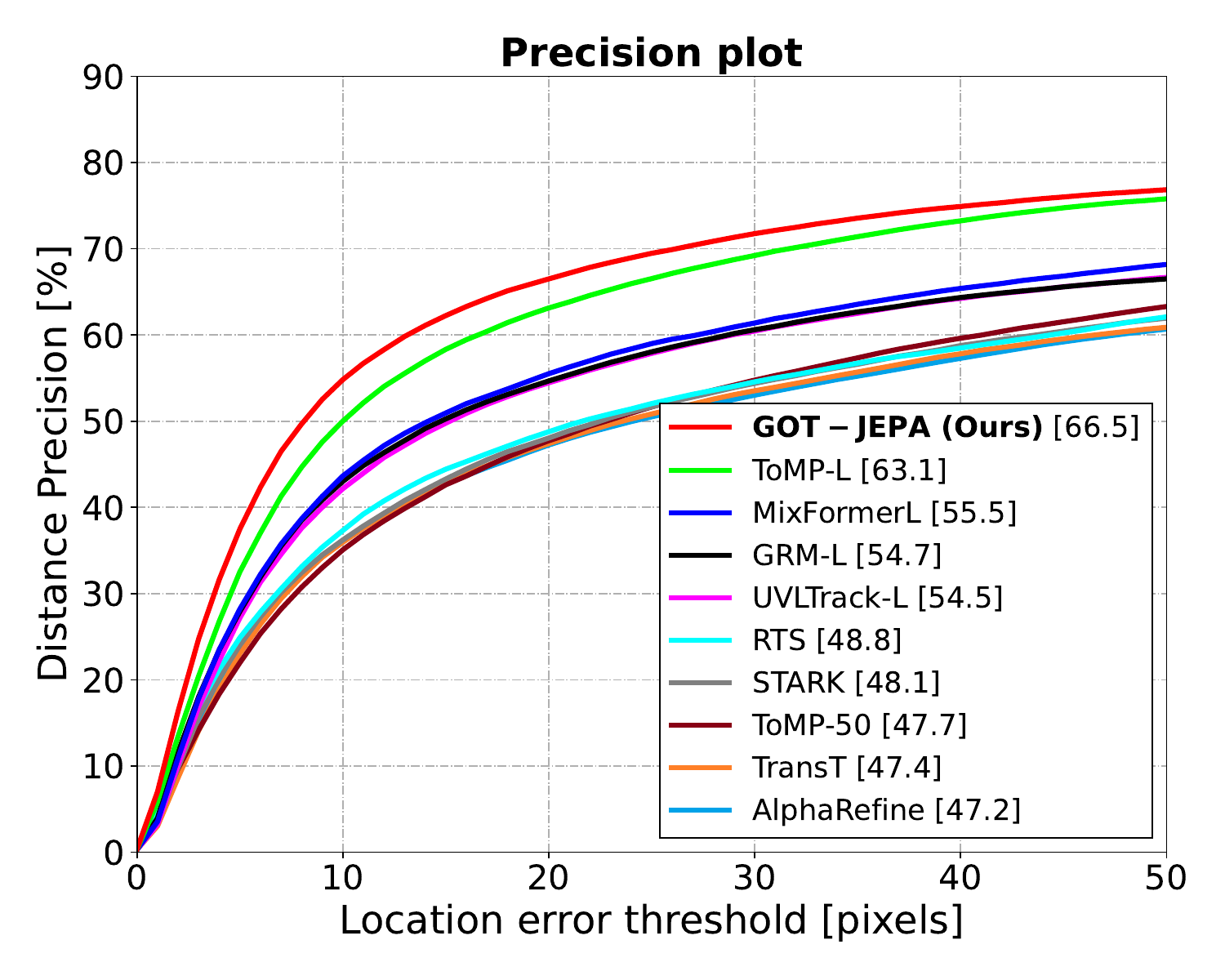}}}&

        \hspace{-0.3cm}
        
        {{\includegraphics[scale=0.22]{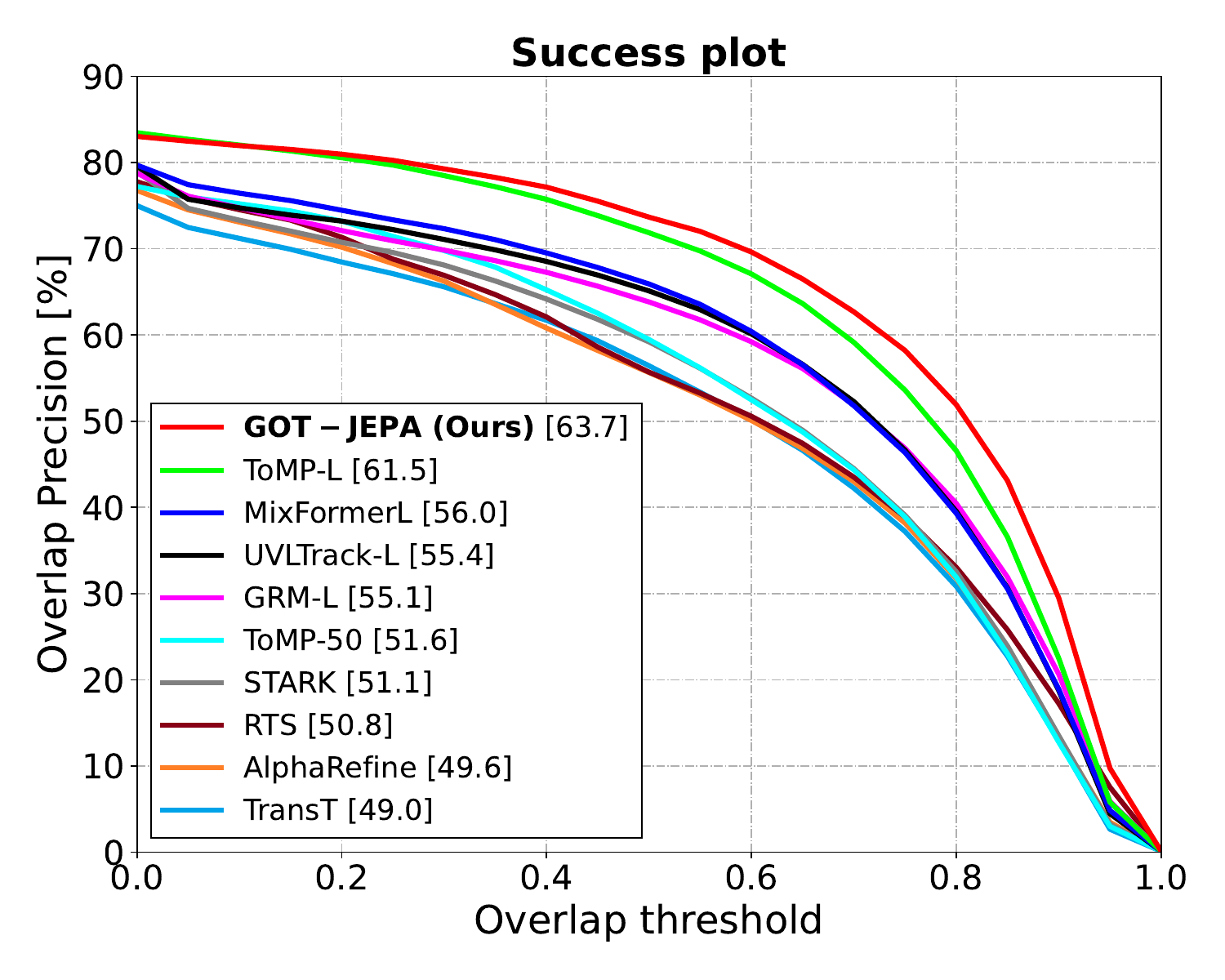}}}

        \end{tabular}
        
	\vspace{-0.15cm}
 
	\caption{
        \textbf{
        Comparison of methods using NPr, Pr, and SUC plots on the AVisT dataset, from left to right.
        } 
        }
	\label{fig:AVisT_all}
\vspace{-0.10in}
\end{figure*}

\noindent \textbf{AVisT}: As shown in~\cref{fig:AVisT_all}, the AVisT dataset, which is training-free and includes diverse adverse scenarios, demonstrates that our tracker consistently surpasses all baselines across metrics and thresholds, even under challenging conditions (NPr $<$ 0.1 or Pr $<$ 10 pixels).

\begin{figure*}[!t]
	\centering
 
        \begin{tabular}{ccc}

        \hspace{-0.3cm}
        
        {{\includegraphics[scale=0.22]{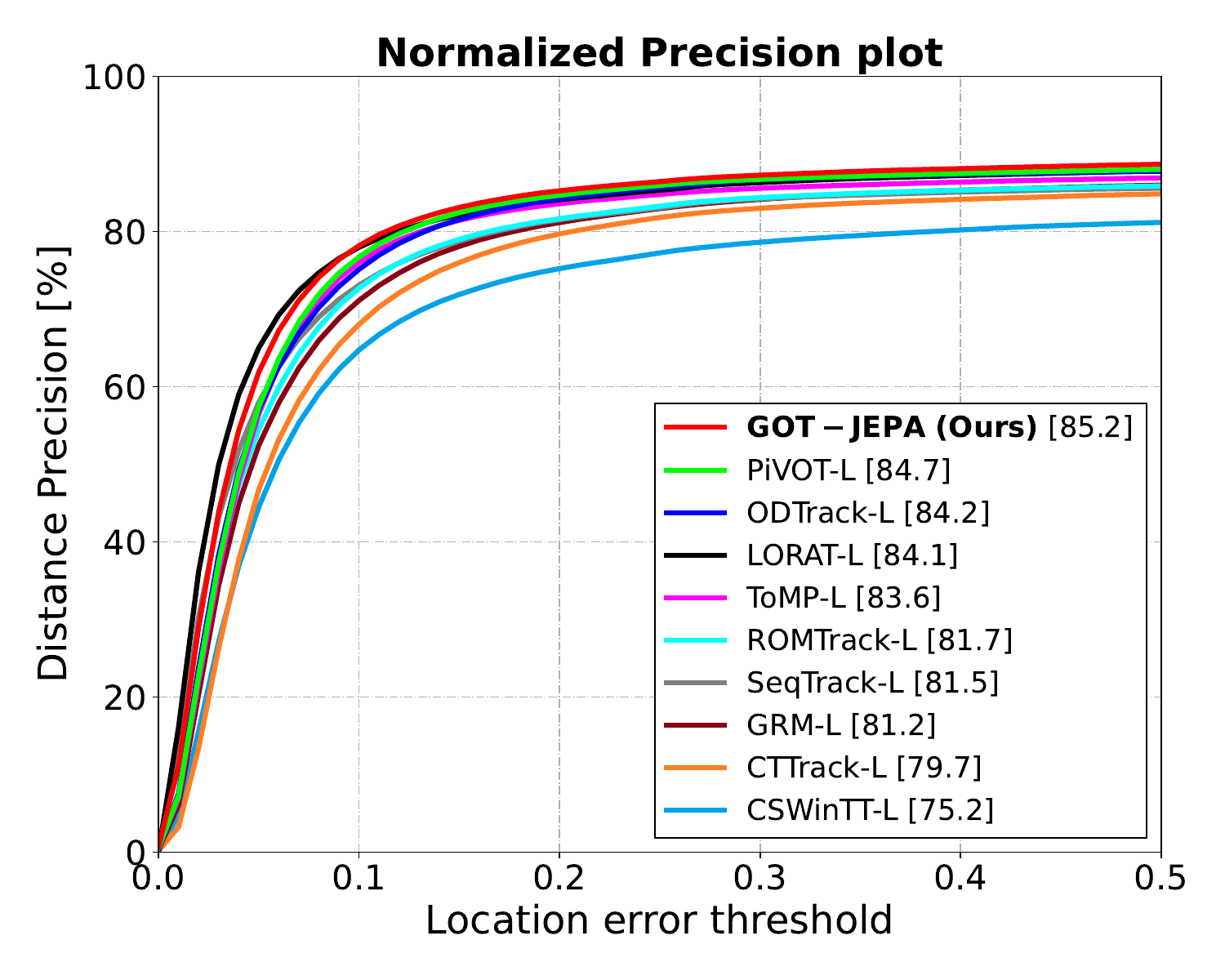}}}&
        
        \hspace{-0.3cm}
        
        {{\includegraphics[scale=0.22]{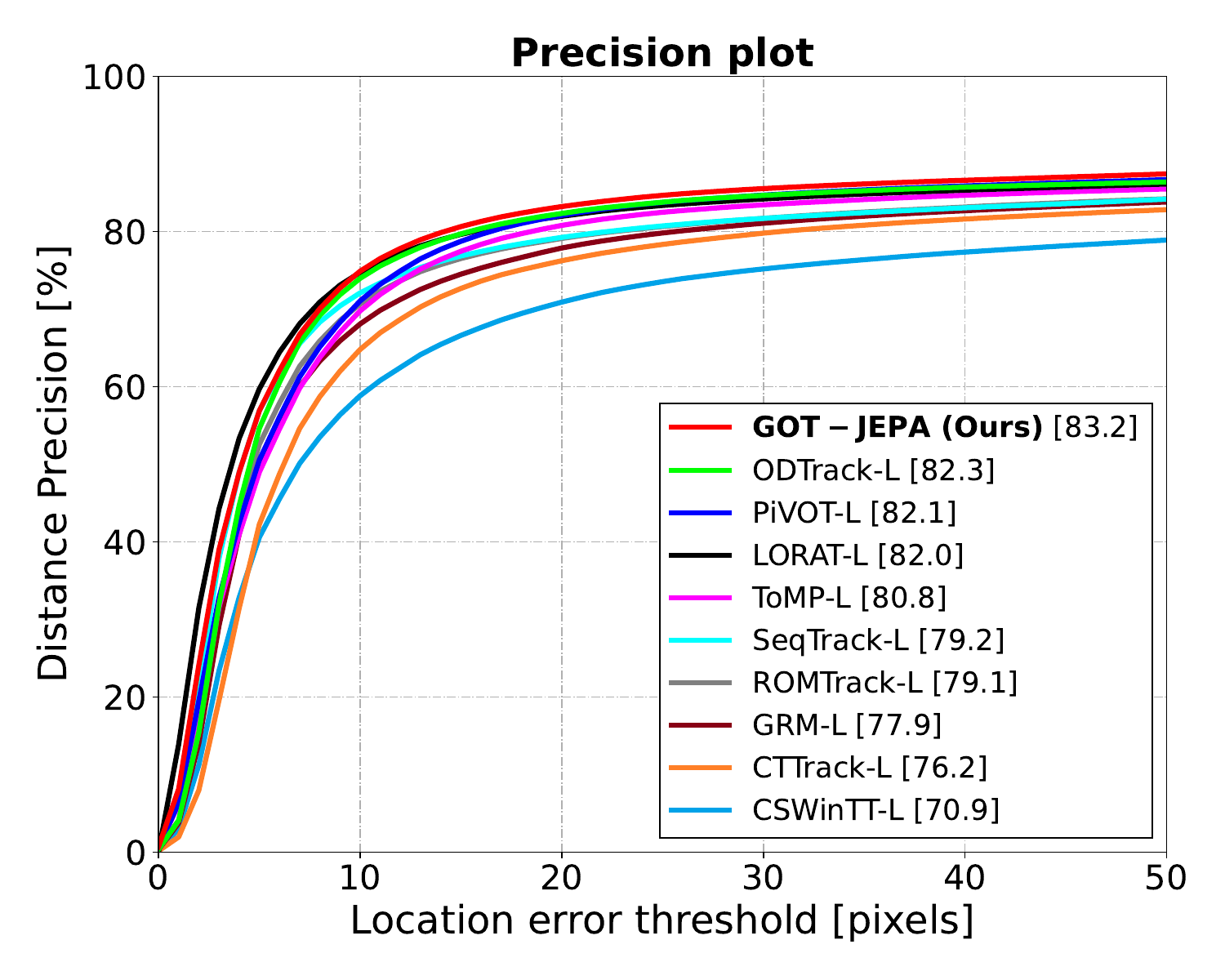}}}&

        \hspace{-0.3cm}
        
        {{\includegraphics[scale=0.22]{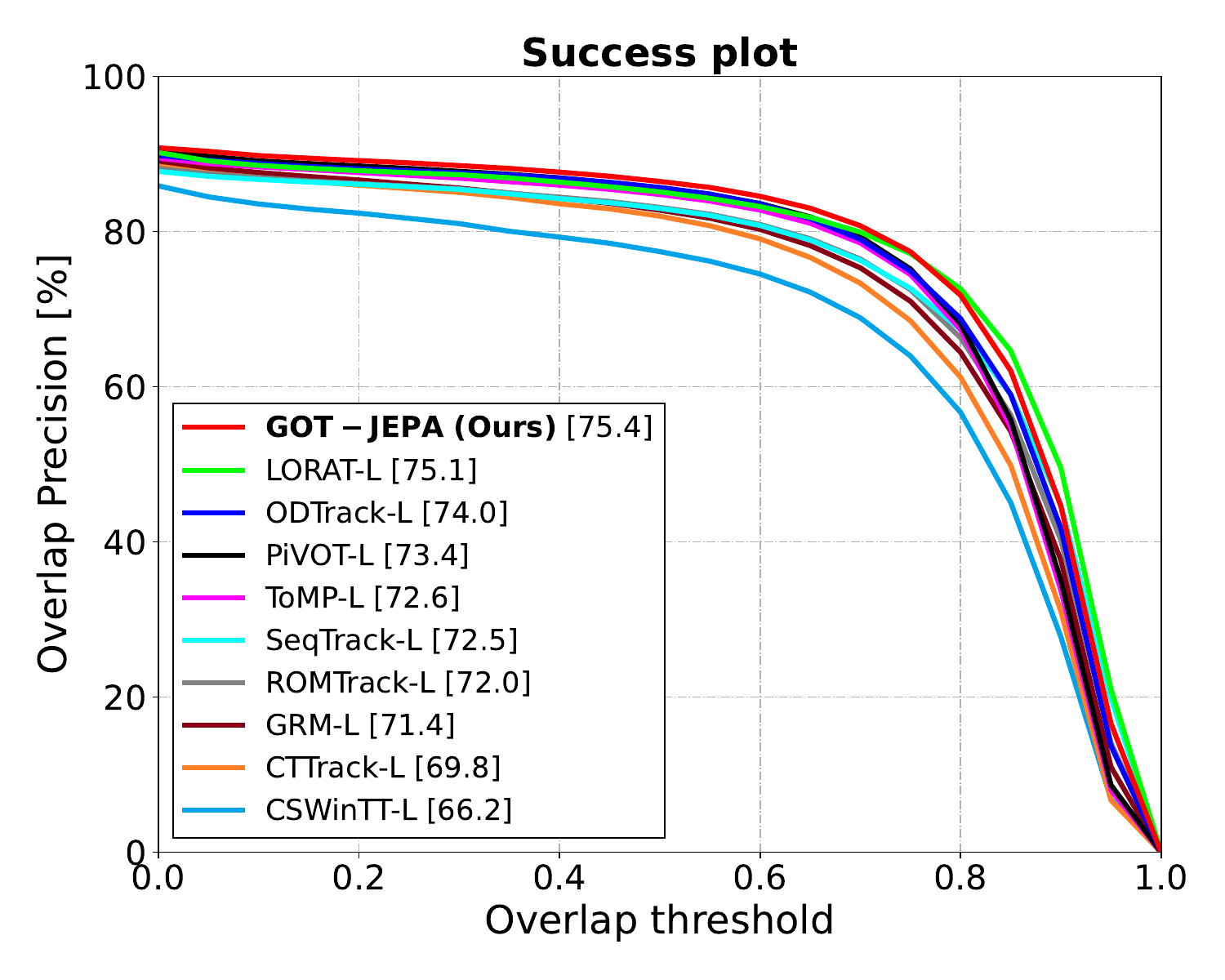}}}
        
        \end{tabular}
        
	\vspace{-0.15cm}
 
	\caption{
        \textbf{
        Comparison of methods using NPr, Pr, and SUC plots on the LaSOT dataset, from left to right.
        } 
        }
	\label{fig:LaSOT_all}
\end{figure*}

\noindent \textbf{LaSOT}: In~\cref{fig:LaSOT_all}, on this in-distribution dataset, our tracker slightly surpasses the state-of-the-art method in Pr and shows a clear advantage in NPr, particularly around the 0.1 threshold, indicating robustness under size-normalized evaluation. 
For SUC, it outperforms most competing trackers once the threshold exceeds 0.6.  





\vspace{-0.1in}
\subsection{Attribute Analysis}
\noindent \textbf{Attribute-Based Analysis:}
We conduct an attribute-based analysis by comparing our method with several state-of-the-art trackers~\cite{UVLTrack,PiVOT,ToMP,GRM,SeqTrack,UPDT,CCOT,ODTrack,EVPTrack,HIPTrack,ROMTrack,ZoomTrack,CTTrack,CSWinTT,SparseTT,MixFormer,UVLTrack,STARK,TransT} using radar plots, as shown in~\cref{fig:radar_plot}. 
This analysis reveals the strengths and weaknesses of different methods and highlights potential areas for improvement.
It is worth noting that attribute-based analysis requires the raw tracking results of each method. 
If the raw data of a tracker is unavailable, or if a dataset lacks an attribute analysis protocol (\eg third-party servers without attribute results), those trackers are excluded from the analysis.

We also provide detailed numeric attribute results for key attributes across three datasets using the large-resolution variants in Table~\ref{tab:attribute_results_large_res}. For ToMP-L~\cite{ToMP,PiVOT}, this is the baseline tracker with a DINOv2 ViT-L backbone. PiVOT~\cite{PiVOT} uses the same baseline tracker as our method but additionally incorporates CLIP~\cite{CLIP} as a semantic feature. ROMTrack~\cite{ROMTrack} and MixFormer~\cite{MixFormer} use different baseline trackers, so we report their results for reference. Since ROMTrack does not report results on AVisT, we include MixFormer results as another variant of a similar method for comparison.

\vspace{0.5em}
\noindent \textbf{OTB-100:} 
As shown in~\cref{fig:radar_plot} (first column), most trackers achieve comparable performance, whereas our tracker exhibits superior overall results.

\vspace{0.5em}
\noindent \textbf{AVisT:} 
As illustrated in~\cref{fig:radar_plot} (second column), our tracker achieves significant improvements across all attributes compared with other trackers, demonstrating its effectiveness in handling unseen and challenging scenarios.

\vspace{0.5em}
\noindent \textbf{LaSOT:} 
As depicted in~\cref{fig:radar_plot} (third column), aided by GOT-JEPA and OccuSolver, our tracker exhibits superior robustness against object deformation, scale variation, and partial invisibility, while maintaining resilience to background clutter when processing in-distribution sequences such as those in LaSOT.

%




\begin{table}[!t]
\caption{Ablation studies on tracker components were evaluated using SUC on three datasets under GOT-JEPA-252.}
\vspace{-0.05in}
\centering
\resizebox{0.42\textwidth}{!}{
\begin{tabular}{c|cc|c|ccc}
\hline
\multicolumn{1}{c|}{} 
& \multicolumn{2}{c|}{JEPA-Pretrain} 
& \multirow{2}{*}{\begin{tabular}[c]{@{}c@{}}OccuSolver\end{tabular}} 
& \multirow{2}{*}{AVisT} 
& \multirow{2}{*}{LaSOT} 
& \multirow{2}{*}{OTB} \\ 
\cline{2-3}
\multicolumn{1}{c|}{} 
& \begin{tabular}[c]{@{}c@{}}Inv. Loss\end{tabular} 
& \begin{tabular}[c]{@{}c@{}}Cov. Loss\end{tabular} 
&  &  &  &  \\
\hline
(1) & - & - & - & 59.2 & 70.7 & 69.2 \\
(2) & - & - & \checkmark & 59.6 & 71.1 & 69.5 \\
(3) & \checkmark & - & - & 60.4 & 72.3 & 70.2 \\
(4) & \checkmark & \checkmark & - & 60.8 & 72.6 & 70.3 \\
(5) & \checkmark & \checkmark & \checkmark & 61.7 & 73.4 & 70.6 \\
\hline
\end{tabular}
\vspace{-0.2in}
}
\label{tab: sALLdataset_tompL_JEPA_Cov_PT}
\end{table}


\begin{table}[ht]
\centering
\color{black}%
\caption{Effect of invariance and covariance loss weights in GOT-JEPA pre-training (L-252, without OccuSolver).}
\label{tab:inv_cov_weights}
\scriptsize
\setlength{\tabcolsep}{3pt}
\renewcommand{\arraystretch}{1.08}
\resizebox{0.3\textwidth}{!}{%
\begin{tabular}{c|cc|cc}
\hline
\begin{tabular}[c]{@{}c@{}}Regularization\\ Method\end{tabular} & $\alpha$ & $\beta$ & AVisT & LaSOT \\
\hline
Inv       & 1  & 0 & 60.4 & 72.3 \\
Inv + Cov & 1  & 1 & 60.1 & 72.1 \\
Inv + Cov & 10 & 1 & 60.6 & 72.5 \\
Inv + Cov & 25 & 1 & 60.8 & 72.6 \\
Inv + Cov & 50 & 1 & 60.7 & 72.6 \\
\hline
\end{tabular}%
}
\end{table}

\begin{table}[!t]
\caption{Ablation studies on the impact of proposed components with varying attributes on the AVisT~\cite{AVisT} dataset.
The bottom row highlights more specific descriptions of the attributes.
}
\vspace{-0.05in}
\resizebox{0.46\textwidth}{!}{
\begin{tabular}{cclc|ccccc}
\hline
\multicolumn{1}{c|}{\multirow{2}{*}{Tracker}} & \multicolumn{2}{c}{\multirow{2}{*}{\begin{tabular}[c]{@{}c@{}}JEPA\\ Pretrain\end{tabular}}} & \multirow{2}{*}{\begin{tabular}[c]{@{}c@{}}Occu\\ Solver\end{tabular}} & \multirow{2}{*}{\begin{tabular}[c]{@{}c@{}}Obstruction \\ Effects\end{tabular}} & \multirow{2}{*}{\begin{tabular}[c]{@{}c@{}}Target \\ Effects\end{tabular}} & \multirow{2}{*}{\begin{tabular}[c]{@{}c@{}}Imaging \\ Effects\end{tabular}} & \multirow{2}{*}{\begin{tabular}[c]{@{}c@{}}Weather \\ Conditions\end{tabular}} & \multirow{2}{*}{Camouflage} \\
\multicolumn{1}{c|}{} & \multicolumn{2}{c}{} &  &  &  &  &  &  \\ \hline
\multicolumn{1}{c|}{ToMP-L} & \multicolumn{2}{c}{-} & - & 56.74 & 42.84 & 54.06 & 62.39 & 65.00 \\
\multicolumn{1}{c|}{GOT-JEPA} & \multicolumn{2}{c}{\checkmark} & - & 58.75 & 46.66 & 54.94 & 66.08 & 62.42 \\
\multicolumn{1}{c|}{GOT-JEPA} & \multicolumn{2}{c}{\checkmark} & \checkmark & 61.86 & 50.59 & 57.44 & 65.83 & 62.45 \\ \hline
\multicolumn{4}{c|}{Attribute Descriptions} & Occlusion & \begin{tabular}[c]{@{}c@{}}Distractor\\ Deformation\end{tabular} & \begin{tabular}[c]{@{}c@{}}Low-light\\ Archival\end{tabular} & \begin{tabular}[c]{@{}c@{}}Target-\\ Visibility\end{tabular} & \begin{tabular}[c]{@{}c@{}}Background-\\ Clutter\end{tabular} \\ \hline
\end{tabular}
}
\label{tab:AVisT_Attr_tompL_252_JEPA_Covariance_PT}
\end{table}



%
\subsection{Ablation Studies}
We conduct several ablation studies in this subsection to validate the effectiveness of each proposed component.
The baselines used in our ablation study include ToMP-L~\cite{PiVOT}, which is the DINOv2~\cite{DINOv2}-based ViT-L variant of ToMP~\cite{ToMP} that utilises the same backbone. Our tracker is built upon the same framework as ToMP-L.

\noindent \textbf{Effect of pre-training, Covariance Loss, and OccuSolver:}
\cref{tab: sALLdataset_tompL_JEPA_Cov_PT} presents the ablation studies on GOT-JEPA, the covariance loss, and OccuSolver across five configurations and three datasets.
The performance difference between the baseline in row~(1) and the full model in row~(5) indicates that GOT-JEPA enhances the ability of the tracker to generate more discriminative features and perform better occlusion reasoning, thereby improving overall tracking performance.
When evaluating OccuSolver alone, a moderate performance gain is observed compared with the baseline tracker (row~(1) vs.\ row~(2)).
However, when OccuSolver receives input from the JEPA-pretrained tracker (row~(4) and row~(5)), the improvement becomes substantially larger.
This is because the JEPA-pretrained tracker provides higher-quality pseudo labels as object priors, enabling OccuSolver to infer more accurate visibility information, which supports generic object tracking.
Moreover, the configuration ``JEPA-Pretrain'' with only the invariance loss (row~(3)) already yields a noticeable improvement over the baseline, and the incorporation of the covariance loss (row~(4)) further enhances the performance by 1.6\%.
Overall, compared with the baseline tracker, the proposed pre-training strategy combined with OccuSolver leads to performance gains of 2.5\% on AVisT, 2.7\% on LaSOT, and 1.4\% on OTB-100.

Table~\ref{tab:inv_cov_weights} reports results for GOT-JEPA-252 without OccuSolver, isolating the effects of the invariance and covariance objectives. The invariance term is necessary for learning from tracking-model predictions, while the covariance term further improves the predictive capability of the tracking models and encourages diverse prediction patterns across frame variants.
The ratio 25:1 provides the best overall trade-off. When the covariance term is overemphasized, as in the 1:1 ratio, performance drops on both AVisT and LaSOT, suggesting that an overly strong covariance objective can hinder invariance-driven prediction learning. When the covariance term is set to a moderate level, such as 10:1 or 50:1, performance is consistently higher than the invariance-only baseline 1:0, indicating that the covariance term is most effective as a regularizer rather than a dominant loss.
Notably, AVisT does not provide training data, so the improvement under ratio-based settings with the covariance loss suggests stronger out-of-distribution generalization. Meanwhile, although LaSOT provides training data, the covariance loss still yields consistent gains, supporting its benefit for the proposed model-adaptation pre-training.

Furthermore, \cref{tab:AVisT_Attr_tompL_252_JEPA_Covariance_PT} analyzes how the proposed components affect individual tracking attributes.
The model predictor, pre-trained with JEPA, enhances the tracker's ability to handle distractors, deformation, and reduced target visibility in adverse weather conditions, such as rain or fog.
OccuSolver further enhances performance in scenarios with occlusion and deformation.

%


\begin{table}[!t]
\centering
\caption{Evaluation of corruption types and their impact on pre-training strategies under GOT-JEPA-252.}
\vspace{-0.05in}
\resizebox{0.38\textwidth}{!}{
\begin{tabular}{cccc|cc}
\hline
 & JEPA-Pretrain & Copy-Paste & Masking & AVisT & LaSOT \\ \hline
(1) &  &  &  & 59.2 & 70.7 \\
(2) &  & \checkmark &  & 59.5 & 71.2 \\
(3) & \checkmark & \checkmark &  & 60.8 & 72.6 \\
(4) & \checkmark &  & \checkmark & 60.0 & 71.3 \\ \hline
\end{tabular}
}
\vspace{-0.1in}
\label{tab: corruption}
\end{table}



\begin{table}[!t]
\centering
\caption{Evaluating the impact of ProjNet on the tracker. GOT-JEPA is evaluated at a resolution of 252 without OccuSolver.}
\vspace{-0.05in}
\resizebox{0.40\textwidth}{!}{
\begin{tabular}{cc|c|ccc}
\hline
\multicolumn{2}{c|}{\multirow{2}{*}{Tracker}} & \multirow{2}{*}{ProjNet} & \multirow{2}{*}{AVisT} & \multirow{2}{*}{LaSOT} & \multirow{2}{*}{OTB-100} \\
\multicolumn{2}{c|}{} &  &  &  &  \\ \hline
(1) & ToMP-L & - & 59.2 & 70.7 & 69.2 \\
(2) & ToMP-L & \checkmark & 58.9 & 70.5 & 69.2 \\
(3) & GOT-JEPA & - & 59.6 & 71.4 & 69.3 \\
(4) & GOT-JEPA & \checkmark & 60.8 & 72.6 & 70.3 \\ \hline
\end{tabular}
}
\label{tab:GOT_JEPA_252_ProjNet_Ablations}
\end{table}


\noindent \textbf{Corruption Types for GOT-JEPA pre-training.}
We employ copy-paste augmentation as the primary corruption technique and also evaluate masking, as reported in~\cref{tab: corruption}. 
Row~(1) shows the ToMP-L baseline at a resolution of 252, while row~(2) introduces copy-paste without JEPA pre-training. 
Rows~(3) and (4) demonstrate that copy-paste achieves larger performance gains when combined with JEPA, although both corruption strategies (\ie copy-paste or masking) yield improvements. 
During pre-training, copy-paste is applied in feature space to the current frame only, thereby simulating distractors and occlusion. 
All experiments in this evaluation are conducted at a resolution of 252 without using OccuSolver.




\begin{figure}[!t] \centering
\includegraphics[width=0.45\textwidth, trim={1.0cm 0cm 2.5cm 1cm}, clip]
{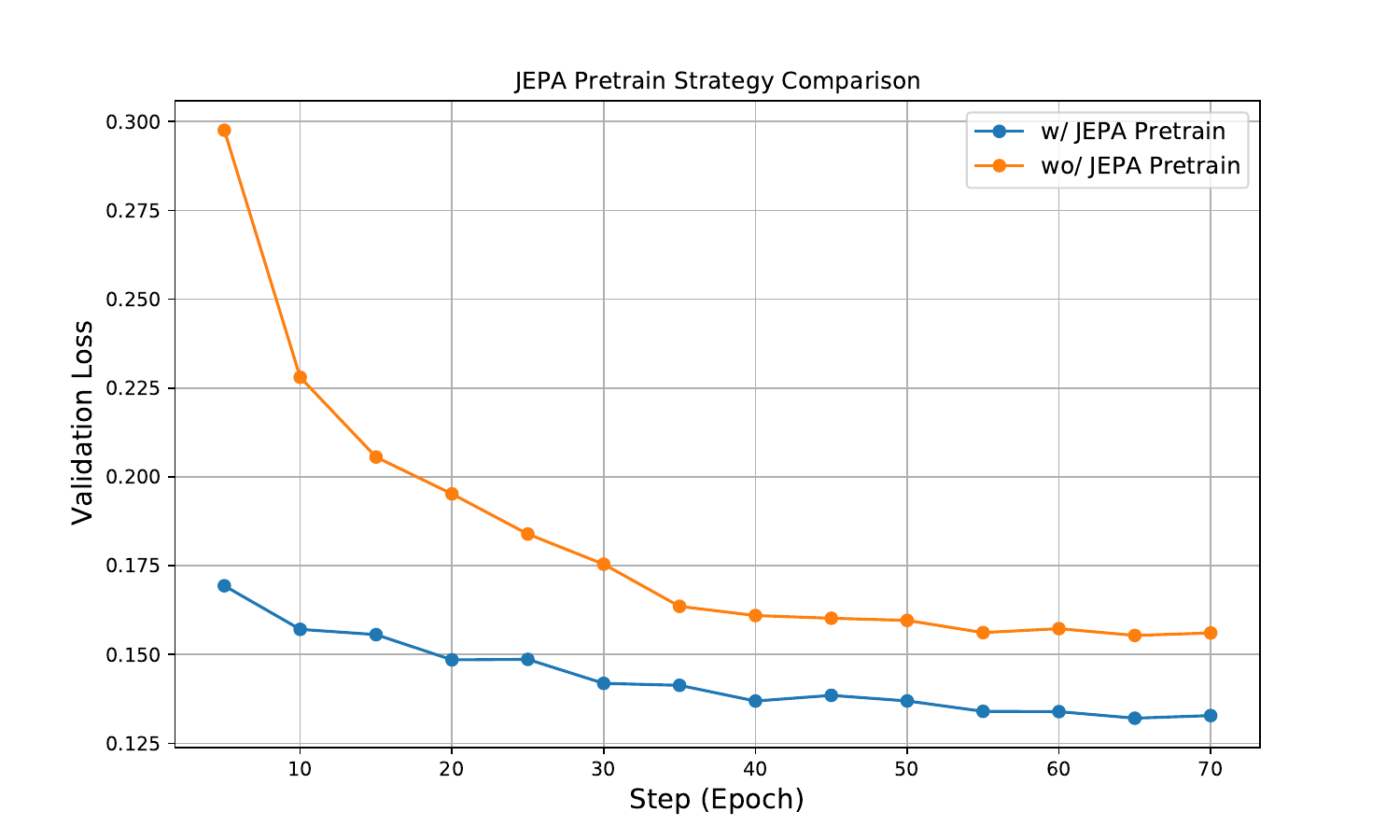}
\vspace{-0.15in}
\caption{
An analysis of the validation curve to study how tracker pre-training, with or without JEPA, affects the learning curve.
} 
\label{fig:JEPA_Pretrain_Comparison}
\end{figure}


\noindent \textbf{Learning Curve Analysis for Tracker Pre-Training from JEPA.}
Besides conducting an ablation study to analyze whether tracker pre-training through the JEPA architecture benefits the tracker, we also provide the learning curve. As indicated in~\cref{fig:JEPA_Pretrain_Comparison}, the tracker using ``JEPA Pretrain,'' specifically validated in training stage 1, learns more effectively than when the JEPA pre-training strategy is not used. 



\begin{table}[!t]
\centering
\caption{The analysis quantifies computational costs of each component in GOT-JEPA-L-378.}
\resizebox{0.48\textwidth}{!}{
\begin{tabular}{c|cc|c|cc|c}
\hline
\multirow{2}{*}{} &
  \multicolumn{2}{c|}{Model Predictor} &
  \multirow{2}{*}{\begin{tabular}[c]{@{}c@{}}Reg/Cls \\ Decoders\end{tabular}} &
  \multicolumn{2}{c|}{OccuSolver} &
  \multirow{2}{*}{Total} \\ \cline{2-3} \cline{5-6}
                        & Backbone & Predictor &      & Point Tracker & Adapters &       \\ \hline
Latency (ms)            & 23.82    & 4.39      & 2.30 & 9.43          & 1.39     & 41.34 \\
Latency (\%)            & 57.65    & 10.6      & 5.6  & 22.8          & 3.4      & 100   \\ \hline
Trainable Parameter (M) & 0        & 17.4      & 2.5  & 0             & 7.7      & 27.6  \\ \hline
{\color{black}MACs (G)}  & {\color{black}125.5} & {\color{black}13.9} & {\color{black}1.8} & {\color{black}182.8} & {\color{black}1.1} & {\color{black}325.1} \\
\hline
\end{tabular}
}
\label{tab:GOT_JEPA_cost}
\end{table}




\noindent \textbf{ProjNet Usability Evaluation.}
We evaluate ProjNet in~\cref{tab:GOT_JEPA_252_ProjNet_Ablations}.
Row~(1) shows the baseline tracker, and row~(2) adds the ProjNet tail to the baseline predictor.
Row~(3) removes ProjNet during both training and inference of GOT-JEPA, while row~(4) includes it.
ProjNet has minimal impact on the baseline tracker (rows~(1) and (2)) due to its lightweight single-layer ($1 \times 1$) convolution design.
Row~(3) confirms that JEPA-based predictor pre-training improves tracking over the baseline, and row~(4) shows further gains when ProjNet is integrated.
ProjNet enhances model prediction learning by projecting the student’s representations into the teacher space, aligning corrupted and clean predictions.

\subsection{Limitations and Future Work}
While GOT-JEPA improves most attributes, as discussed in Section IV.C, its limitations are present in scenarios involving dense background clutter and fast motion, as evidenced by performance on the LaSOT and OTB-100 benchmarks. Furthermore, out-of-distribution data, such as those in AVisT, remains an open challenge for further improvement, especially under adverse imaging conditions (e.g., low-quality frames) and extreme target conditions (e.g., tiny objects moving rapidly).
Looking forward, a promising direction to address background clutter is to incorporate 3D geometric cues, as most current trackers primarily rely on 2D semantic features. This can be achieved by augmenting training with auxiliary modalities (e.g., RGB-X data) or by leveraging recent advances in 3D geometry research~\cite{gotedit2026iclr,VGGT} to predict geometric representations from 2D images. Such geometry-aware cues may significantly enhance robustness in complex environments by more effectively separating the target from cluttered backgrounds.



\begin{table}[!t]
\centering
\caption{Comparison of different query points for OccuSolver AUC as a metric across three datasets.}
\resizebox{0.30\textwidth}{!}{
\begin{tabular}{c|ccc}
\hline
{\#Points} & {AVisT} & {LaSOT} & {OTB-100} \\ \hline
64 & 61.2 & 72.9 & 70.3 \\
128 & 61.7 & 73.3 & 70.6 \\
256 & 61.8 & 73.3 & 70.4 \\ \hline
\end{tabular}
}
\vspace{-0.15in}
\label{tab:point_num}
\end{table}

\begin{table}[!t]
\centering
\caption{An ablation study on whether preventing point sampling from the occluded frame can benefit tracking.}
\resizebox{0.4\textwidth}{!}{
\begin{tabular}{c|ccc}
\hline
First Frame Constraint & AVisT & LaSOT & OTB-100 \\ \hline
- & 63.55 & 75.03 & 73.24 \\
\checkmark  & 63.69 & 75.36 & 73.24 \\
\hline
\end{tabular}
}

\label{tab:OccuSolver_first_frame_constraint}
\end{table}

\begin{table}[!t]
\centering
\caption{Comparison between using the original Point Tracker for OccuSolver and our refinement strategy on the AVisT.
}
\resizebox{0.48\textwidth}{!}{
\begin{tabular}{c|ccc|ccc}
\hline
Tracker                   & w/ Point Tracker & w/ LS-FineTune & w/ Obj. Prior & NPr  & SUC  & OP50 \\ \hline
\multirow{3}{*}{GOT-JEPA} & \checkmark       &                &               & 81.2 & 63.0 & 73.2 \\
                          & \checkmark       & \checkmark     &               & 81.3 & 63.3 & 73.3 \\
                          & \checkmark       & \checkmark     & \checkmark    & 81.9 & 63.7 & 73.7 \\ \hline
\end{tabular}
}
\label{tab:point_refinement}
\end{table}

%

\vspace{0.1in}
\section{Computational Cost Analysis}

\cref{tab:GOT_JEPA_cost} presents the per-frame runtime (ms), percentage of total time, trainable parameters, and multiply–accumulate operations (MACs) for each tracker component.
With high-resolution inputs ($378 \times 378$), the Model Predictor uses DINO features to estimate the tracking model, while the Adapters in OccuSolver refine the Point Tracker output, and the Reg/Cls decoders generate the final predictions.
The primary computational bottleneck is the Backbone, which employs ViT-L at a large spatial resolution.
The Point Tracker also incurs a notable cost, as it processes multiple query points.
Since the cost scales linearly with the number of points~\cite{Cotracker}, we fix the number of query points to 128 for efficiency, compared with dense tracking tasks that process millions of points (see~\cref{tab:point_num}).

\begin{figure}[!t] \centering
\includegraphics[width=0.48\textwidth]{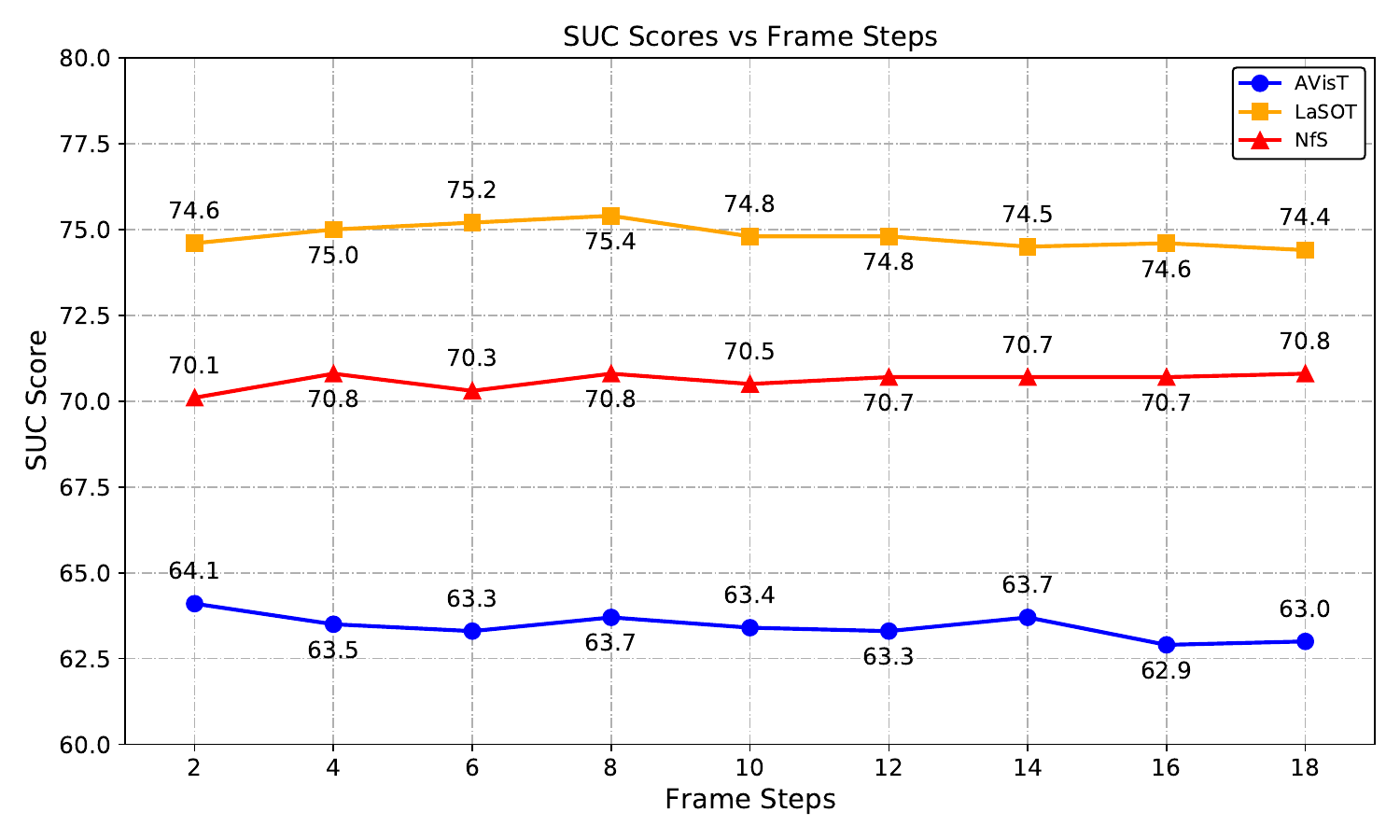}
%
\caption{
An ablation study investigates the frame gap between consecutive sampled frames for OccuSolver during inference, using the SUC as the metric.
} 
\label{fig:FM_T-Frame_step_PT}
%
\end{figure}

\noindent \textbf{Frame Sampling Strategy of OccuSolver:} 
\cref{fig:FM_T-Frame_step_PT} examines the effect of the sampling step (N) between consecutive frames processed by OccuSolver. Since OccuSolver uses a fixed number of input frames, (N) defines a trade-off. A small (N) improves performance on moderate motion but weakens robustness to long occlusions, whereas a large (N) increases tracking failures under significant motion. Our results show (N=8) achieves the best balance.

To enhance stability, we add an inference constraint that prevents initialization on heavily occluded targets. If the first-frame visibility score drops below a threshold (e.g., lower than 85\% visible points), initialization is skipped. During tracking, the input window follows a FIFO scheme; when a new frame is predicted occluded, the last unoccluded frame is duplicated. As shown in \cref{tab:OccuSolver_first_frame_constraint}, this yields consistent gains (up to 0.3). We apply this constraint by default.


\begin{figure*}[!ht]
\centering

\includegraphics[width=0.98\linewidth, 
trim={0cm 0 0cm 0}, clip]{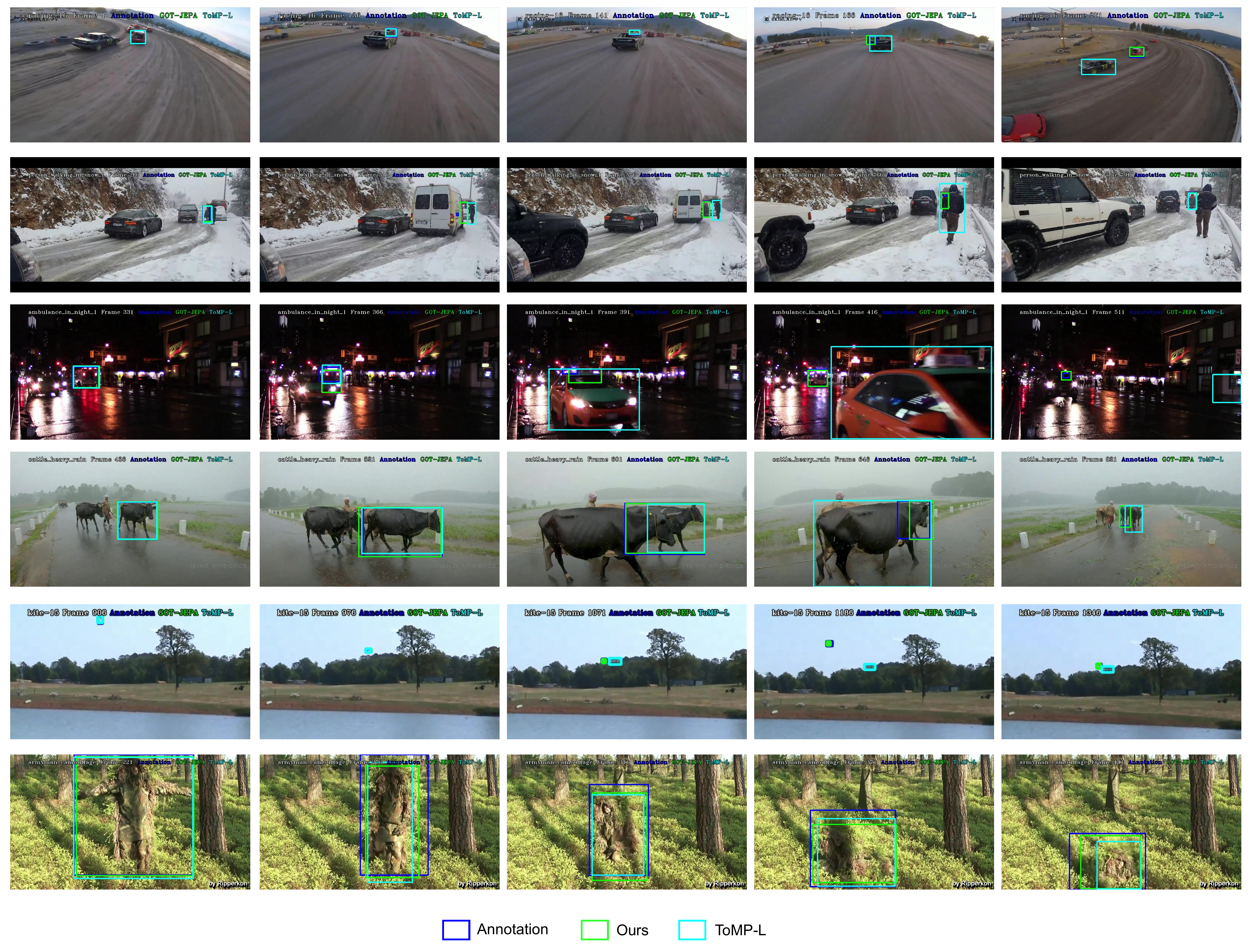}

%
\vspace{-0.3 cm}

\caption{
Visual comparisons of tracking results from raw annotations, GOT-JEPA, and the baseline tracker ToMP-L across diverse video sequences under adverse scenarios are presented.
Rows 1 to 3 illustrate object tracking under occlusion, including partial and full occlusion, and the tracker's recovery after occlusion removal.
Row 4 shows occlusion under distractor cases. Row 5 shows tracking with small targets. Row 6 presents tracking results for deformation cases.
Best viewed when zoomed in; more detailed visual comparisons among trackers are provided in the video appendix.
}
\label{fig:vis_res}
\end{figure*}


\vspace{-0.2in}


\begin{figure}[!ht] \centering
\includegraphics[width=0.45\textwidth, trim={1.0cm 0cm 2.5cm 1cm}, clip]
{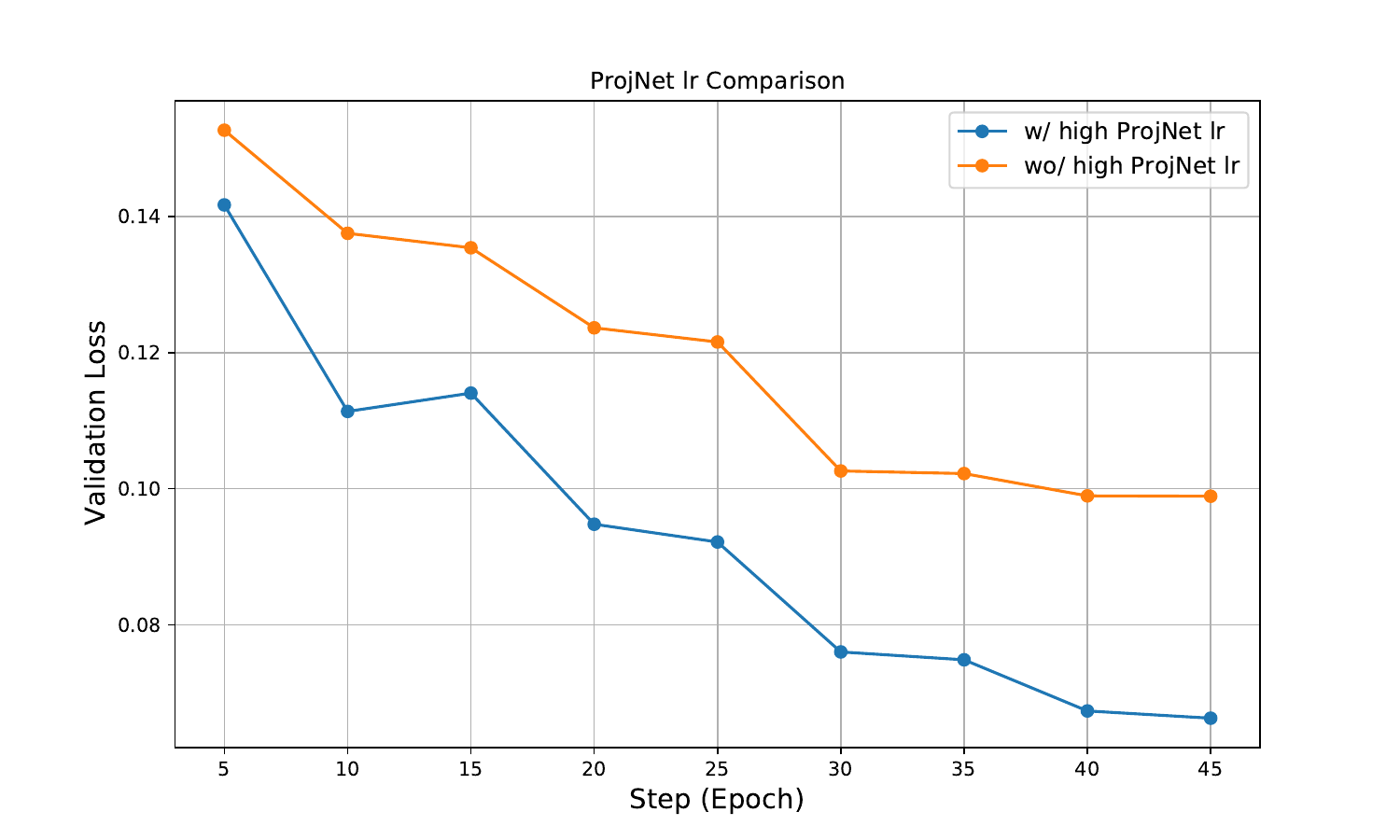}
\vspace{-0.15in}
\caption{
Comparison of relative learning rates between ProjNet and other components during the JEPA training stage.
} 
\label{fig:ProjNet_lr_Comparison}
\end{figure}


\begin{figure}[!ht] \centering
%
\includegraphics[width=0.47 \textwidth, trim={0cm 0cm 0cm 0cm}, clip]{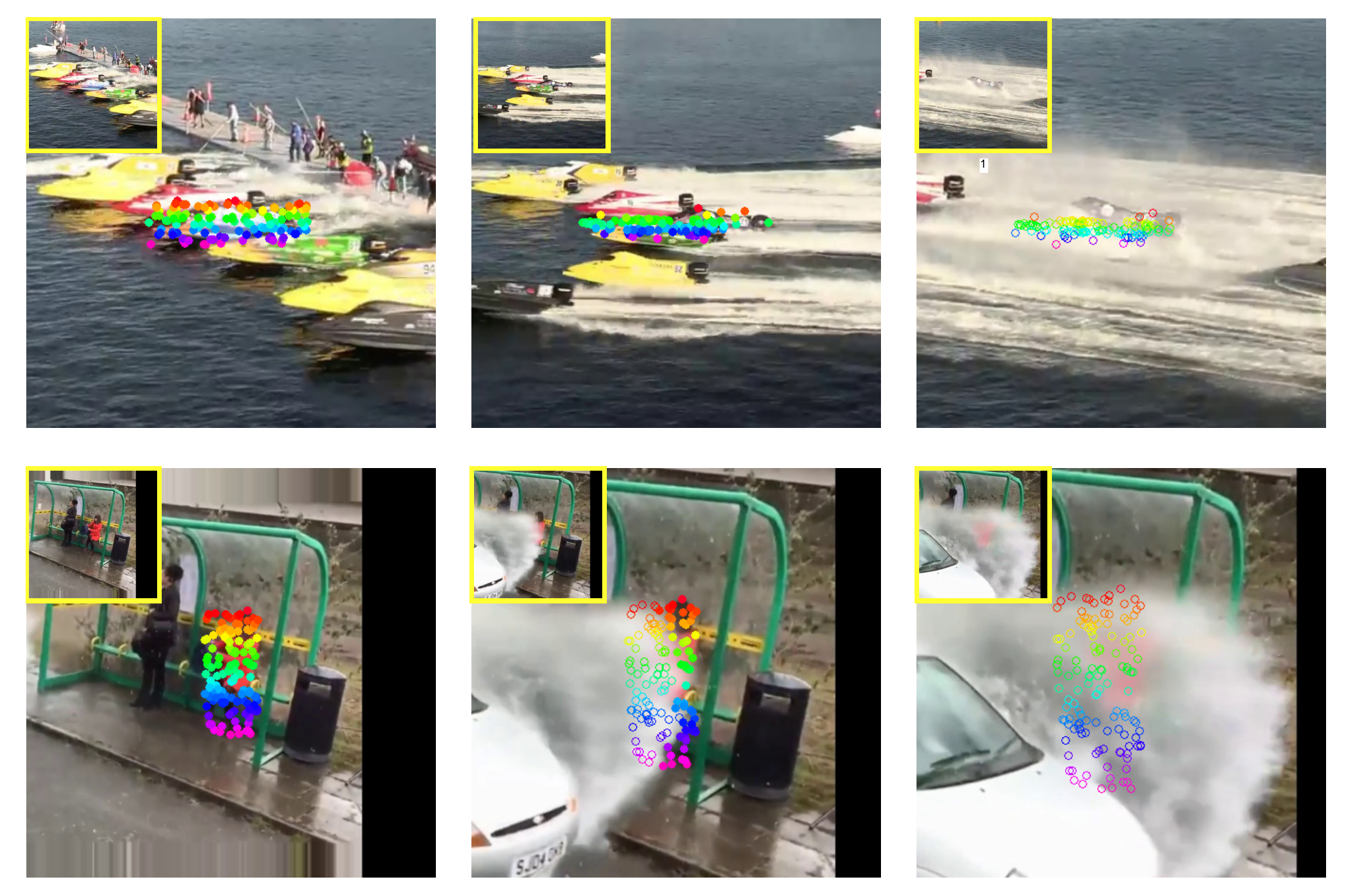}
\vspace{-0.10in}
\caption{Col 1: Initial frame and initial point sampling.  
Col 2: Points that are more object-aware and can handle partial occlusion.  
Col 3: Cases of full occlusion.
The original input image is shown in the yellow box.
} 
\label{fig:point_vis_got_jepa}
\vspace{-0.1in}
\end{figure}


\vspace{0.1in}
\noindent \textbf{Refinement of Point Tracker.}  
We provide an ablation study in \cref{tab:point_refinement} to quantify the impact of different point-tracker (CoTracker) variants in GOT-JEPA. Each row uses GOT-JEPA as the base tracker. The first setting uses only the pretrained point tracker and its predicted visibility, which is converted by our proposed mapping function for GOT feature refinement, without ladder-side fine-tuning or object priors. The second setting adds ladder-side fine-tuning of the point tracker, and the third setting further incorporates GOT-derived object priors.
As shown in \cref{tab:point_refinement}, ladder-side fine-tuning of the point tracker and GOT-derived object priors yields consistent improvements, and the full setting achieves the best performance.
We further clarify that CoTracker alone is not an occlusion solution for GOT. CoTracker is class-agnostic and outputs point trajectories without target identity and without target-specific query initialization, so it does not provide object-aware visibility on its own. In our framework, GOT-derived object priors and ladder-side fine-tuning make the point trajectories target-centric and align them with the GOT tracking feature space through our proposed mapping function, producing object-aware visibility cues that the tracker can directly exploit.


\vspace{0.1in}
\noindent \textbf{Relative Learning Rate between ProjNet and Other Components.}
In~\cref{fig:ProjNet_lr_Comparison}, we analyze the performance difference in the relative learning rate between ProjNet and other components (such as the Expander and Student model predictor).
A higher learning rate for ProjNet enhances the model's learning effectiveness. The final learning rate for ProjNet is ten times that of the other components, starting at 1e-3 and 1e-4, respectively. This experiment was conducted with an image resolution of 252 during the model predictor pre-training stage, focusing on learning to predict tracking models.
Similar phenomena can be observed in previous works with JEPA-like structure, such as DirectPred~\cite{DirectPred}.


\vspace{-0.1in}
\subsection{Visualization Results}
%
We present visual comparisons among trackers in~\cref{fig:vis_res}. Our tracker achieves more robust tracking under occlusion (rows 1 to 3) and superior discrimination and robustness against distractors, small targets, and deformation (rows 4 to 6). These advantages stem from our proposed model-based prediction learning and OccuSolver, which enhance the tracker's ability to handle occlusions and resist distractors.

We also present OccuSolver's output in~\cref{fig:point_vis_got_jepa}, illustrating how randomly initialized points are refined to enhance object awareness or handle occlusions. Each column shows the original image alongside the point sampling, followed by the prediction result: Col 1 presents the initial frame and point sampling; Col 2 shows points that are more object-aware and handle partial occlusion; Col 3 presents cases of full occlusion. More qualitative results are in the supplementary material.


\section{Conclusion}
\label{sec:conclusion}

In this work, we introduced GOT-JEPA, a tracking framework that enhances online model prediction and adaptation by advancing the JEPA from image feature prediction to the novel task of tracking model prediction. A teacher predictor generates pseudo-tracking models, while a student predictor learns to predict these models from corrupted current frames, given identical previous frame information. This design empowers the student model to both learn from diverse tracking concepts and account for potential variations, allowing it to predict an adapted and generalized model for the current frame. Equipped with OccuSolver, GOT-JEPA further enhances occlusion perception by generating precise, object-aware visibility states. This information not only improves occlusion handling but also produces higher-quality reference labels that incrementally refine and enhance subsequent model predictions, thereby boosting the tracker's overall robustness and adaptability in complex environments.


\bibliographystyle{IEEEtran}
\bibliography{main}



\section*{Biography Section}
\vspace{-13 mm}

\begin{IEEEbiography}[{\includegraphics[width=1in,height=1.25in,clip,keepaspectratio]{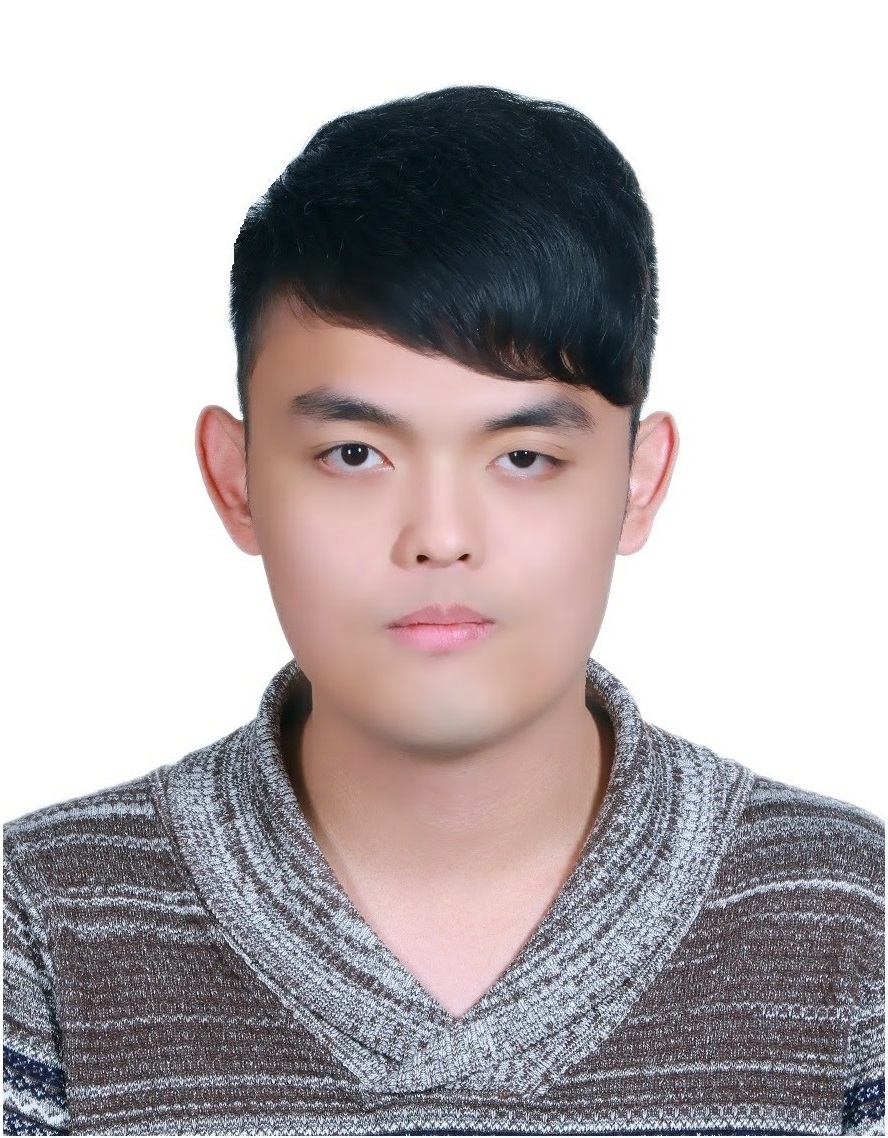}}]
{Shih-Fang Chen} is a Ph.D. candidate in the Department of Computer Science at National Yang Ming Chiao Tung University, Taiwan. His research interests primarily focus on computer vision and machine learning. 
He actively contributes to the research field as a reviewer for several leading international journals and conferences, including IEEE Transactions on Pattern Analysis and Machine Intelligence (TPAMI), IEEE Transactions on Circuits and Systems for Video Technology (TCSVT), IEEE Transactions on Multimedia (TMM), and the IEEE/CVF Conference on Computer Vision and Pattern Recognition (CVPR).
\end{IEEEbiography}

\vspace{-12 mm}

\begin{IEEEbiography}
[{\includegraphics[width=1in,height=1.25in,clip,keepaspectratio] {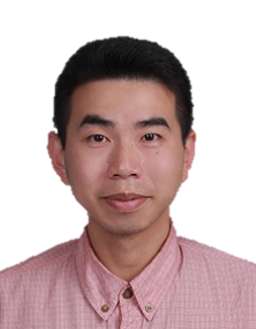}}]
{Jun-Cheng Chen} (Member, IEEE) is an Associate Research Fellow at the Research Center for Information Technology Innovation (CITI), Academia Sinica. He joined CITI as an assistant research fellow in 2019. He received the BS and MS degrees advised by Prof. Ja-Ling Wu in Computer Science and Information Engineering from National Taiwan University, Taiwan (R.O.C), in 2004 and 2006, respectively, where he received the PhD degree advised by Prof. Rama Chellappa in Computer Science from University of Maryland, College Park, USA, in 2016. From 2017 to 2019, he was a postdoctoral research fellow at the University of Maryland Institute for Advanced Computer Studies. His research interests include computer vision, machine learning, deep learning and their applications to biometrics, such as face recognition/facial analytics, activity recognition/detection in the visual surveillance domain, etc. His works have been recognized in prestigious journals and conferences in the field, including PNAS, TBIOM, CVPR, ICCV, ECCV, ICLR, WACV, etc. He was a recipient of the ACM Multimedia Best Technical Full Paper Award in 2006, APSIPA ASC Best Paper Award in 2023, and IEEE CE Magazine Best Paper Award in 2025.
\end{IEEEbiography}

\vspace{-12 mm}

\begin{IEEEbiography}
[{\includegraphics[width=1in,height=1.25in,clip,keepaspectratio] {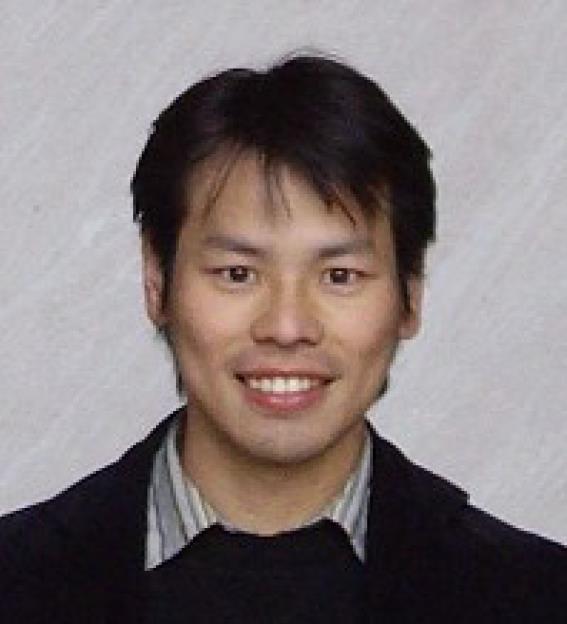}}]
{I-Hong Jhuo}  (Member, IEEE) is a Principal Applied Scientist at Microsoft AI. He actively contributes to the development of innovative technologies for computer vision, information retrieval and large-scale data systems. His research interests include computer vision, information retrieval and artificial intelligence. His work has been recognized in international competitions, including the ACM Multimedia Grand Challenge 2012. He also contributed to the design of a top-performing video analytics system for the NIST TRECVID Multimedia Event Detection (MED) evaluation in collaboration with Columbia University, New York, NY, USA.
\end{IEEEbiography}

\vspace{-12 mm}

\begin{IEEEbiography}[{\includegraphics[width=1in,height=1.25in,clip,keepaspectratio]{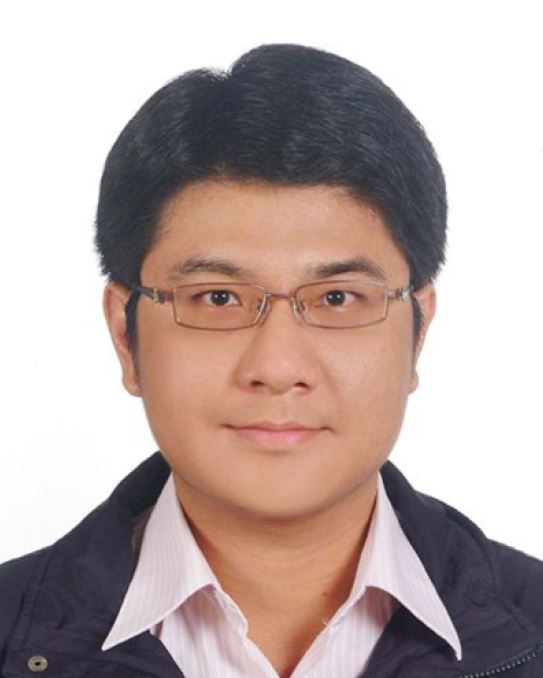}}]%
{Yen-Yu Lin} (Senior Member, IEEE) 
received the B.B.A. degree in Information Management, and the M.S. and Ph.D. degrees in Computer Science and Information Engineering from National Taiwan University, Taipei, Taiwan, in 2001, 2003, and 2010, respectively. He is currently a Distinguished Professor with the Department of Computer Science, National Yang Ming Chiao Tung University, Hsinchu, Taiwan. His research interests include computer vision, machine learning, and artificial intelligence. He has served as an Area Chair for ICCV, CVPR, and ECCV. He is currently an Associate Editor of IEEE Transactions on Pattern Analysis and Machine Intelligence (TPAMI) and the International Journal of Computer Vision (IJCV).
\end{IEEEbiography}









\end{document}